%% file: main.tex
\documentclass[lettersize,journal]{IEEEtran}

\usepackage{cite}

\usepackage{amsmath}
\usepackage{amssymb}
\usepackage{gensymb}

\usepackage{algorithm}
\usepackage{algpseudocode}
\algrenewcommand\algorithmicrequire{\textbf{Input:}}
\algrenewcommand\algorithmicensure{\textbf{Output:}}
\algrenewcommand\algorithmicindent{1em}

\usepackage{booktabs}
\usepackage{array}
\usepackage{rotating}
\usepackage{textcomp}
\usepackage{stfloats}
\usepackage{url}
\usepackage{verbatim}
\usepackage{graphicx}
\usepackage{subcaption}
\usepackage{multirow}

\usepackage{threeparttable}
\usepackage[colorlinks=true, linkcolor=blue, citecolor=blue, urlcolor=blue]{hyperref}
\usepackage{color}

\begin{document}

\title{Unified Dual-Decoder Deep Reinforcement Learning for Close-Enough Traveling Salesman Problem}

\author{Mingfeng Fan*, Jiaqi Cheng*, Yaoxin Wu, Yifeng Zhang\textsuperscript{$\dag$}, Yibin Yang, Guohua Wu, and Guillaume Sartoretti
\thanks{Mingfeng Fan and Jiaqi Cheng contributed equally to this work.}
\thanks{Mingfeng Fan, Yifeng Zhang, and Guillaume Sartoretti are with the Department of Mechanical Engineering, National University of Singapore, Singapore (E-mail: ming.fan@nus.edu.sg, yifeng@u.nus.edu, guillaume.sartoretti@nus.edu.sg).}
\thanks{Jiaqi Cheng is with Meituan, China (e-mail: chengjq0705@gmail.com).}
\thanks{Yaoxin Wu is with the Department of Industrial Engineering and Innovation Sciences, Eindhoven University of Technology, Netherlands. (E-mail: wyxacc@hotmail.com)}
\thanks{Yibin Yang is with the School of Vehicle and Mobility, Tsinghua University, China. (E-mail: yangybtiecun15@gmail.com)}
\thanks{Guohua Wu is with the School of Automation, Central South University, China. (E-mail: guohuawu@csu.edu.cn)}
\thanks{Corresponding authors: Yifeng Zhang.}
}



\maketitle

\begin{abstract}
Many transportation and logistics tasks, such as automatic meter reading, drone-based inspection, and sensor data collection, require a vehicle to pass close enough to target regions rather than visit exact locations. These tasks are modeled as the close-enough traveling salesman problem (CETSP), where both the visiting sequence and the waypoint within each neighborhood must be determined. Existing deep reinforcement learning (DRL) solvers target standard routing problems with fixed node locations and do not directly handle this coupled node--waypoint decision. We formulate a Markov decision process for CETSP with a hybrid discrete--continuous action space and propose a unified dual-decoder DRL framework, termed UD3RL. The node-decoder selects the next target neighborhood, and the loc-decoder outputs a von Mises distribution over its boundary angle, so that the waypoint lies anywhere on the boundary. A geometry-aware $k$-nearest-neighbor subgraph interaction mechanism encodes the relative geometry of nearby neighborhoods in a local frame aligned with the arrival direction and masks unreachable neighbors to enhance local spatial reasoning for waypoint selection. A customized REINFORCE algorithm trains one model across problem sizes and radius configurations. Extensive experiments show that UD3RL achieves a favorable trade-off between solution quality and computational efficiency, while generalizing well across problem scales and distributions. With lightweight refinement, it achieves performance competitive with or better than the state-of-the-art CETSP solver and effectively handles dynamic requests without retraining.


\end{abstract}


\section{Introduction}
\label{sec:introduction}

\begin{figure}[t]
\centering
\vspace{-5mm}
\includegraphics[width=0.8\columnwidth]{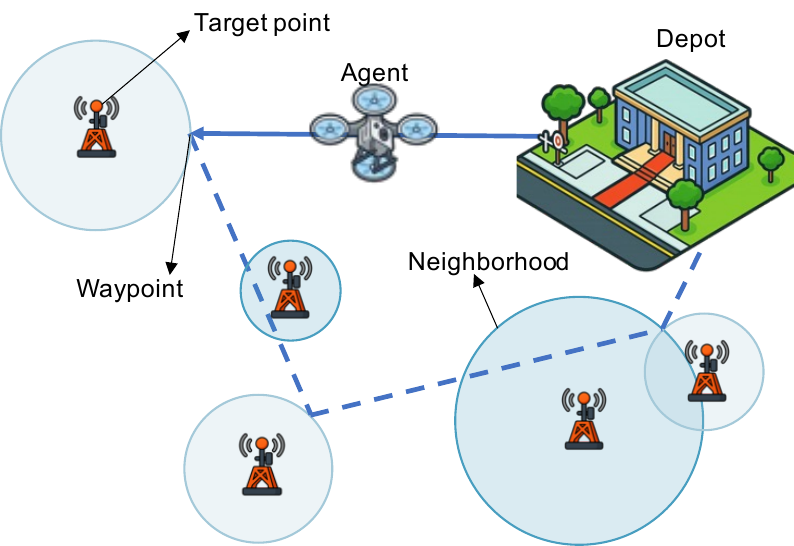}
\caption{An example path in the CETSP, where an agent is required to visit a set of target neighborhoods, each represented by a circular region. }
\label{fig:sketch}
\vspace{-3mm}
\end{figure}

\IEEEPARstart{G}{round} vehicles and unmanned aerial vehicles (UAVs) are increasingly deployed in transportation and logistics systems to perform inspection, sensing, data collection, and delivery services~\cite{coutinho2018unmanned}. In many such tasks, a vehicle does not need to visit the exact coordinate of each target. Instead, the task is completed once the vehicle enters a sensing, communication, or service region around the target. Examples include automatic meter reading, in which a utility vehicle driving through a street network reads radio-frequency identification (RFID) meters within a given radius~\cite{gulczynski2006close, shuttleworth2008advances}, data collection from wireless sensor networks by a mobile robot that only needs to enter the communication range of each sensor~\cite{yuan2007optimal}, robot path planning~\cite{shin2022deep}, and diagnostic reconnaissance of photovoltaic panels by UAVs~\cite{di2022genetic}. These applications can be naturally modeled as the close-enough traveling salesman problem (CETSP), a variant of the traveling salesman problem (TSP). In the CETSP, each target is associated with a neighborhood, typically represented by a circular region centered at the target, and the vehicle only needs to pass through one feasible point within each neighborhood~\cite{carrabs2017novel}, as shown in Fig.~\ref{fig:sketch}.

Compared with the classical TSP, CETSP is more challenging because it involves two coupled decisions: selecting the order of target neighborhoods and determining the specific waypoint used to visit each selected neighborhood. The former is a discrete combinatorial decision, while the latter is a continuous geometric decision~\cite{{lei2024effective}}. These two decisions are strongly coupled, since a different waypoint choice can change the preferred visiting order, and a different visiting order can change the optimal waypoint inside each neighborhood. 

Many approaches have been proposed in the literature to solve the CETSP, which can generally be categorized into exact and heuristic algorithms. Exact algorithms~\cite{behdani2014integer,coutinho2016branch} are capable of providing optimal solutions for moderately sized instances, but they suffer from scalability issues and are impractical for large-scale problems. Heuristic methods~\cite{mennell2009heuristics, lei2024effective}, while able to handle larger instances, tend to become computationally expensive as the problem size increases, and often rely heavily on domain-specific knowledge. These limitations are particularly restrictive in transportation and logistics operations, where routes must be generated or updated quickly as new requests arrive, as in on-demand delivery and dynamic pickup-and-delivery services~\cite{zhou2020two, li2021learning}.

In recent years, deep reinforcement learning (DRL) has emerged as a promising alternative, attracting considerable attention for addressing combinatorial optimization problems (COPs), with notable examples including AM~\cite{kool2018attention}, POMO~\cite{kwon2020pomo}, and LEHD~\cite{luo2023neural}. Compared to traditional methods, DRL offers lower inference costs and greater flexibility across different problem domains. Despite these advantages, DRL has rarely been applied to the CETSP. The main difficulty lies in the waypoint selection subtask after the node visiting sequence is determined. This subtask can be formulated as a second-order cone programming (SOCP) problem, and its solution may in turn affect the preferred node sequence. Such bidirectional dependency makes the CETSP challenging for existing DRL-based routing methods.

To overcome this challenge, we exploit the fact that the waypoints of an optimal CETSP tour lie on the boundaries of the target neighborhoods~\cite{behdani2014integer}, and represent each waypoint by its boundary angle. On this basis, we propose a \underline{u}nified \underline{d}ual-\underline{d}ecoder \underline{DRL} framework, termed UD3RL, which solves the CETSP directly by decoupling the decision-making process into a discrete node selection and a continuous waypoint selection. Specifically, UD3RL employs an adapted Transformer encoder to extract problem representations and utilizes a node-decoder and a loc-decoder to sequentially select the next node to visit and the precise location (i.e., waypoint) on the boundary of its neighborhood. The loc-decoder outputs a von Mises distribution over the boundary angle, allowing continuous waypoint selection without discretizing the neighborhood. Since each waypoint is selected together with its node, the length of a constructed tour is available without solving an SOCP. UD3RL is therefore trained end-to-end on this length, so that the visiting order is optimized for the close-enough tour rather than for the tour through the target centers. When new requests appear, it re-plans the remaining route from the vehicle's current position without calling a solver.
Our contributions are summarized as follows.
\begin{enumerate}
    \item We propose a flexible architecture, UD3RL, for solving both static and dynamic CETSP instances. It features a node-decoder for selecting the next node to visit and a loc-decoder that outputs a von Mises distribution over the boundary angle of the chosen neighborhood, which forms a hybrid discrete--continuous policy without discretizing the neighborhoods.
    \item We introduce a geometry-aware $k$-nearest neighbors ($k$-NN) subgraph interaction strategy for the loc-decoder. It describes the $k$ nearest neighborhoods by their relative geometry in a local frame aligned with the arrival direction and masks unreachable neighbors, which guides the DRL agent toward short waypoint choices.
    \item We conduct extensive experiments on static, generalized, benchmark, and dynamic CETSP instances. The results demonstrate that UD3RL achieves competitive solution quality with high inference efficiency, generalizes well across problem scales and distributions, and effectively adapts to dynamic requests without retraining.
\end{enumerate}

The remainder of this paper is organized as follows: Section~\ref{sec:related} reviews traditional methods for solving the CETSP and DRL approaches for TSP. Section~\ref{sec:preliminaries} presents the problem definition and Markov Decision Process (MDP) model for CETSP. Section~\ref{sec:method} details the proposed UD3RL framework. Section~\ref{sec:exp} reports and analyzes the experimental results. Finally, Section~\ref{sec:conclude} concludes the paper and discusses potential directions for future work.

\section{Related Works} \label{sec:related}
In this section, we review conventional methods for solving the CETSP as well as DRL approaches for the TSP. To date, relatively few studies have applied DRL to CETSP.

\subsection{Conventional Methods for CETSP} 
The CETSP was first addressed in~\cite{gulczynski2006close}, which proposed a three-phase heuristic consisting of: (1) selecting a feasible set of supernodes $\mathcal{G}$, (2) generating a feasible tour $\tau$ over the points in $\mathcal{G}$, and (3) improving the tour $\tau$. Following this framework, subsequent works~\cite{mennell2009heuristics, mennell2011steiner, wang2019steiner} adopted similar three-step heuristics based on the concept of Steiner zones.
Exact algorithms include mixed-integer programming (MIP) approaches based on discretization schemes~\cite{behdani2014integer, carrabs2017improved, carrabs2017novel} and a branch-and-bound algorithm utilizing SOCP~\cite{coutinho2016branch}. While MIP-based methods offer flexible modeling, their solution quality heavily depends on the discretization granularity. The SOCP-based approach can yield optimal solutions for moderately sized instances but does not scale well to larger problems.
Building on~\cite{carrabs2017improved}, Carrabs et al.\cite{carrabs2020adaptive} proposed a meta-heuristic method incorporating an adaptive discretization scheme. Other heuristic methods\cite{di2022genetic, lei2024effective} directly tackle CETSP using meta-heuristics combined with SOCP formulations. However, such meta-heuristics often require substantial expert knowledge and become computationally expensive as problem size increases, limiting their practical scalability. Faigl~\cite{faigl2018gsoa} proposed the growing self-organizing array (GSOA), an unsupervised learning procedure that adapts an array of nodes to the target neighborhoods with low computational requirements.

\subsection{DRL Approaches for TSP} 
DRL models can learn effective patterns from large datasets without relying on handcrafted heuristics or domain-specific knowledge, enabling strong generalization to unseen instances~\cite{DRL-Searcher}. Moreover, once trained, DRL models offer high inference efficiency. These advantages have led to a surge of research applying DRL to solve the classical TSP in recent years.
Based on their solution generation strategies, DRL-based routing methods can be broadly classified into two categories: \emph{improvement-based} approaches~\cite{chen2019learning, hottung2022neural, wu2021learning, d2020learning, kim2021learning, ma2021learning} and \emph{construction-based} approaches~\cite{bello2016neural, nazari2018reinforcement, kool2018attention, xin2021multi, kwon2020pomo, kim2022sym}. Among these, construction-based methods generally exhibit faster inference times~\cite{li2022overview, berto2025rl4co}, making them particularly suitable for solving the CETSP.
Specifically, Bello et al.\cite{bello2016neural} proposed one of the earliest DRL-based methods for the TSP by training a pointer network (PtrNet)\cite{vinyals2015pointer} using an actor-critic framework. Subsequently, Kool et al.\cite{kool2018attention} introduced the attention model (AM), which replaced PtrNet with a Transformer-based architecture\cite{vaswani2017attention} to achieve higher solution quality. Building on this, POMO~\cite{kwon2020pomo} further improved performance by leveraging multiple parallel solution trajectories and applying data augmentation to address the symmetry in routing problems, achieving near-optimal results for the TSP. More recent works~\cite{luo2023neural, li2023t2t} integrate supervised learning to further improve performance. However, due to the complexity of CETSP, these DRL-based methods are not directly applicable for solving CETSP.

Two further lines of work relate to the design of UD3RL. First, several DRL methods handle hybrid action spaces in which a discrete choice is paired with continuous parameters, either by alternating between learning the discrete choice and its parameters~\cite{masson2016reinforcement} or by extending deterministic policy gradients to parameterized actions~\cite{hausknecht2016deep}. These methods address control tasks with a fixed set of discrete actions, whereas the discrete part of a CETSP action is a target selected from a varying set of unvisited targets and its continuous part is an angle on a circle. UD3RL therefore pairs an attention-based node-decoder with a von Mises distribution, the circular analogue of the normal distribution~\cite{mardia2000directional}, for the waypoint angle. Second, recent neural routing solvers improve generalization by restricting the decoder to a local view of the $k$ nearest nodes, such as the transferrable local policy of ELG~\cite{gao2024towards} and the invariant nested views of INViT~\cite{fang2024invit}, while Sym-NCO~\cite{kim2022sym} exploits the symmetry of routing problems through its training objective. These methods select nodes in standard routing problems. UD3RL instead uses a local $k$-NN subgraph to select waypoints, describing each neighbor by its geometry relative to the selected neighborhood in a frame aligned with the travel direction.

\section{Preliminaries} \label{sec:preliminaries}

\subsection{Problem Definition and Notation}
Let there be \( n+1 \) nodes located in a two-dimensional plane, indexed by the set \( N = \{0, 1, \dots, n\} \), and let index \( 0 \) denote the depot and indexes in $\{1,2\cdots,n\}$ as the \textit{target points}.  
Each node  $i \in N$ is associated with a closed disk $C_i$ centered at $o_i=\{cx_i, cy_i\}$ with radius $r_i$ where $r_0=0$. We define the \textit{neighborhood} of node $i$ as $\mathcal{N}(i) := C_i$ where $C_0=o_0$. 
The objective of the CETSP is to find the shortest tour $\tau^*$ that starts and ends at the depot $0$ and intersects the neighborhood \( \mathcal{N}(i) \) for every target point, in any order. Let $\{y_0, y_1, y_2, \dots, y_g\}$ where $y_0=o_0$ be the set of \textit{waypoints} in the tour $\tau$, i.e., points at which the tour changes direction. Any tour $\tau$ can be uniquely identified by its sequence of waypoints. 
The cost of an edge $y_iy_{i+1}$ is defined as the Euclidean distance between $y_i$ and $y_{i+1}$. The total cost of a tour $\tau$ is denoted as $L(\tau)$, which is the sum of the edge lengths: $L(\tau) = \sum_{i=0}^{g-1}\|y_{i+1} - y_i\| + \|y_0-y_g\|$.
The goal is to find an optimal tour $\tau^*$ such that $\tau^* = \arg\min_{\tau} L(\tau)$ subject to the constraints that \( \tau \) forms a cycle starting and ending at the depot \( o_0 \), and intersects all neighborhoods \( \mathcal{N}(i) \) for all target points.

\subsection{Waypoint Parameterization}
Since each neighborhood \( \mathcal{N}(i) \), \( i \in \{1, \cdots, n\} \), contains an infinite number of possible waypoints, the number of feasible tours for the CETSP is also infinite. However, the waypoints of the optimal tour \( \tau^* \) are always located on the boundaries of the neighborhoods~\cite{behdani2014integer}. We therefore represent a waypoint selected in \( \mathcal{N}(i) \) by an angle \( \psi \in [-\pi, \pi) \) on the boundary of \( \mathcal{N}(i) \):
\begin{equation}\label{eq:waypoint}
    y(i, \psi) = o_i + r_i\,[\cos\psi,\ \sin\psi].
\end{equation}
Every angle yields a feasible waypoint, and the angles cover the entire boundary of \( \mathcal{N}(i) \). Note that it is not necessary to select a waypoint in every neighborhood, as some edges of a tour can pass through certain neighborhoods without placing a waypoint within them. Discretization schemes such as the \textit{Perimetral Discretization Scheme} (PDS)~\cite{carrabs2017novel} instead restrict each neighborhood to a fixed set of \( \gamma \) candidate points, so that the solution quality depends on the discretization granularity. The parameterization in Eq.~(\ref{eq:waypoint}) keeps one scalar decision per neighborhood and introduces no discretization error. The PDS-based discrete variant of UD3RL, which we compare with in Section~\ref{subsec:cont_vs_disc}, is described in Appendix~\ref{app:pds}.

\subsection{MDP for CETSP}
The route construction in CETSP can be viewed as a sequential decision-making process, which can be naturally modeled as an MDP and solved using reinforcement learning (RL). We define the elements of the MDP, including the state space, action space, transition rule, and reward function, as follows.

\noindent\textbf{State:} The state at time step \( t \), denoted by \( S_t = \left \{\pi_t, Z_t \right \} \), consists of the current partial solution \( \pi_t \) and the set of already visited target points \( Z_t \).
The partial solution \( \pi_t = \left \{X_t, \tau_t \right \} \) includes:
\begin{itemize}
    \item The sequence of selected nodes up to time step \( t \), denoted by \( X_t = \{x_0, x_1, \dots, x_t\} \), where \( x_0 \) is the depot and \( x_j \in N \).
    \item The corresponding sequence of selected waypoints \( \tau_t = \{y_0, y_1, \dots, y_t\} \), where \( y_j = y(x_j, \psi_j) \) lies on the boundary of \( \mathcal{N}(x_j) \) for \( j > 0 \), and \( y_0 = o_0 \) is the depot coordinate.
\end{itemize}
The visited set \( Z_t = \bigcup_{j=1}^{t} z_j \), where each \( z_j \subseteq N \) contains the set of target points whose neighborhoods are intersected by the edge \( y_{j-1}y_j \) at time step \( j \).

\noindent\textbf{Action:} At each time step \( t \), the agent selects an action \( a_t = (x_{t+1}, \psi_{t+1}) \), where \( x_{t+1} \in N \setminus Z_t \) is the next node to be visited, and \( \psi_{t+1} \in [-\pi, \pi) \) is the boundary angle that determines the waypoint \( y_{t+1} = y(x_{t+1}, \psi_{t+1}) \) according to Eq.~(\ref{eq:waypoint}). The action space is therefore hybrid, pairing a discrete node choice with a continuous angle. The process continues until the terminal state is reached, i.e., when the agent returns to the depot and completes the tour.

\noindent\textbf{Transition:} The next state \( S_{t+1} = \left(\pi_{t+1}, Z_{t+1}\right) \) is determined by the selected action \( a_t = (x_{t+1}, \psi_{t+1} ) \) at time step \( t \). The partial solution is updated by appending the new node and waypoint to the current state, i.e., $\pi_{t+1} = \left(X_{t+1}, \tau_{t+1}\right) = \left(X_t \cup \{x_{t+1}\},\ \tau_t \cup \{y_{t+1}\}\right)$.
If \( x_{t+1} = 0 \) (i.e., the depot), then \( y_{t+1} = o_0 \) since \( r_0 = 0 \), and the process transitions to the terminal state. The set of visited nodes is updated by merging \( Z_t \) with the set of newly passed nodes \( z_{t+1} \), whose neighborhoods are intersected by the edge \( y_t y_{t+1} \), i.e., $Z_{t+1} = Z_t \cup z_{t+1}$.

\noindent\textbf{Reward:} To encourage shorter tours, the reward is defined as the negative of the total travel distance. The cumulative reward for a complete tour \( \tau = \{y_0, y_1, \cdots, y_T\} \) is given by: $R = -L(\tau)$, where \( L(\tau) \) is the total Euclidean length of the tour constructed from the sequence of selected waypoints.

\section{Methodology} \label{sec:method}
In this section, we introduce the proposed UD3RL (unified dual-decoder DRL), including its architecture and training algorithm.

\subsection{Overview}
\begin{figure*}[t]
\centering
\vspace{-5mm}
\includegraphics[width=\textwidth]{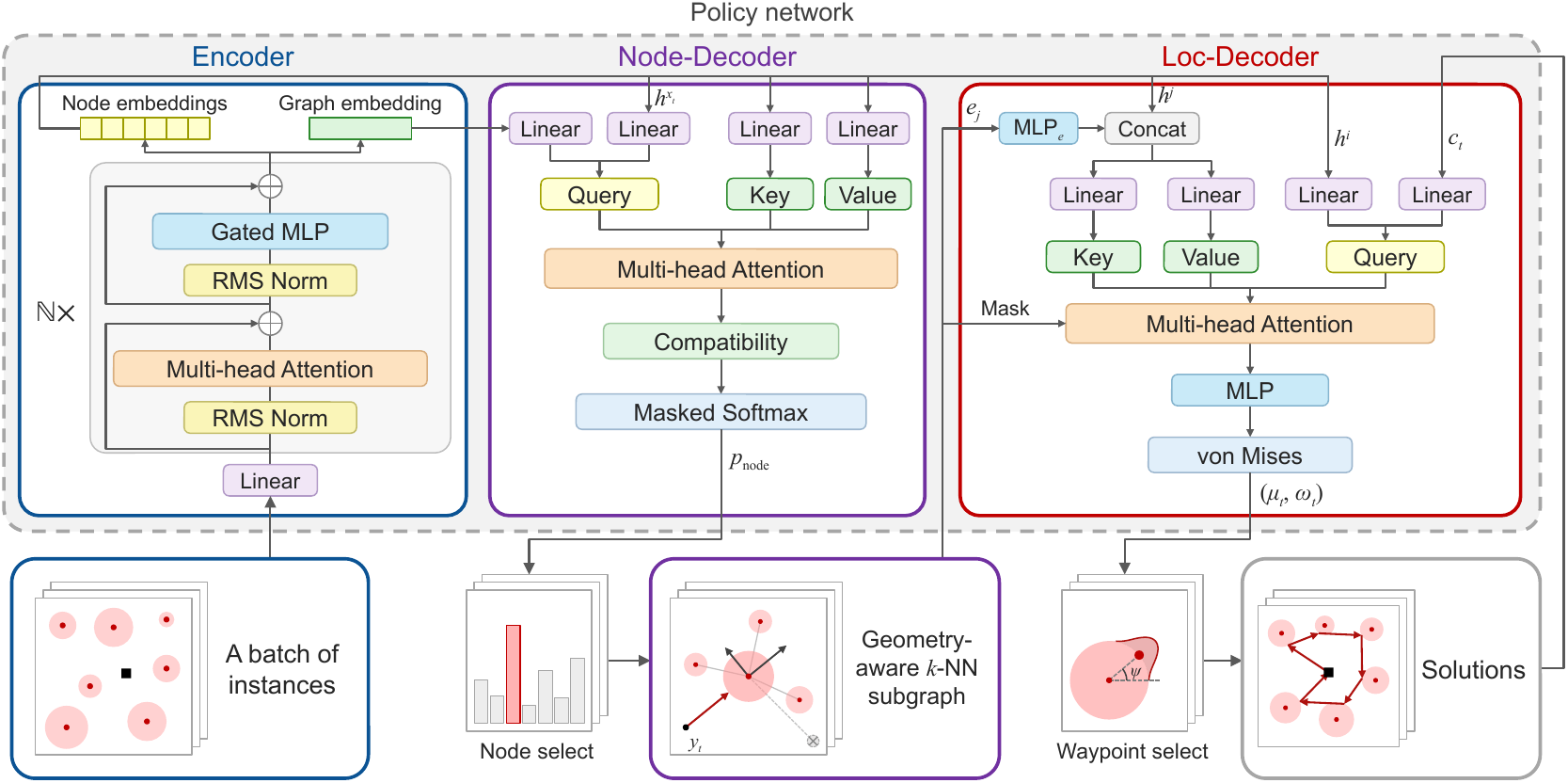}
\caption{Architecture of the policy network used in our method. It consists of an adapted Transformer encoder, a node-decoder, and a loc-decoder. Given a batch of instances, the encoder generates node embeddings and a graph embedding. Based on these, the node-decoder outputs selection probabilities for the next node to visit, while the loc-decoder outputs the mean and concentration of a von Mises distribution over the boundary angle of the waypoint in the selected neighborhood. The loc-decoder attends to a geometry-aware $k$-NN subgraph, in which each neighbor is described by its node embedding and its relative geometry in a local frame aligned with the arrival direction, and unreachable neighbors are masked.}
\label{network}
\vspace{-3mm}
\end{figure*}

As illustrated in Fig.~\ref{network}, the UD3RL policy network $p_\theta$ with trainable parameters $\theta$ is tailored to address the CETSP and consists of an adapted Transformer encoder, a node-decoder, and a loc-decoder. The encoder processes the input node features, which include the depot location, target locations, and the radii of their respective neighborhoods, and maps them into feature embeddings, thereby extracting essential information from the data. Given the feature embeddings and problem-specific constraints, the node-decoder selects the next node to visit. Subsequently, the loc-decoder determines the waypoint on the boundary of the chosen node's neighborhood through a von Mises distribution over the boundary angle. The two decoders sequentially select actions at each time step to construct a complete solution.

The final output of the policy network \( p_\theta \) is a sequence of action tuples \( (x_t, \psi_t) \), consisting of selected nodes and the boundary angles of their waypoints. The output solution is denoted as $\pi = \left\{(x_0, \psi_0), (x_1, \psi_1), \dots, (x_{T}, \psi_{T})\right\}, x_t \in N,\ \psi_t \in [-\pi, \pi)$,
where \( T \) is the time step limits for solution construction. The probability density of generating a solution \( \pi \) is defined by the learned policy \( p_\theta \) as:
\begin{equation}
    p_\theta\left(\pi \mid s\right) = \prod_{t=1}^{T} p_\theta^{\mathrm{node}}\left(x_{t} \mid s, \pi_{t-1}\right) f_\theta^{\mathrm{loc}}\left(\psi_{t} \mid s, \pi_{t-1}, x_t\right),
\end{equation}
where \( s \) represents the current CETSP instance, \( \pi_{t-1} \) is the partial solution up to time \( t-1 \), \( p_\theta^{\mathrm{node}} \) is the categorical distribution over nodes produced by the node-decoder, and \( f_\theta^{\mathrm{loc}} \) is the von Mises density over boundary angles produced by the loc-decoder.

\subsection{Transformer Encoder}
The encoder is an adaptation of the AM~\cite{kool2018attention}, consisting of a stack of attention layers. Each layer includes a multi-head attention (MHA) sublayer and a feed-forward (FF) sublayer, enhanced by several modifications. Building on~\cite{berto2024routefinder}, we replace the original ReLU-activated MLP with a SiLU-activated gated MLP~\cite{shazeer2020glu} in the FF sublayer, enhancing representational power. Additionally, we apply root mean square (RMS) normalization~\cite{zhang2019root} in place of layer normalization and adopt a pre-norm structure (i.e., applying normalization before the residual connections) in each sublayer to improve training stability and accelerate convergence.
The detailed operations of the attention layer are provided in Appendix~\ref{app:attention}.
 
\noindent\textbf{Raw Features to Hidden States.} 
We use the node coordinates \(\{o_i\}_{i=0}^n \) and their corresponding neighborhood radii \( \{r_i\}_{i=0}^n \) as raw input features. We project the target point coordinates and neighborhood radii into their respective embeddings using linear transformations:
\begin{equation}
    h_{o,0}^i = W_o o_i, \quad h_{r,0}^i = W_r r_i, \quad \forall i \in N,
\end{equation}
where \( W_o \) and \( W_r \) are learnable projection matrices.
The initial node embeddings are then formed by concatenating the coordinate and radius embeddings: $h_0^i = [h_{o,0}^i;\ h_{r,0}^i], \quad \forall i \in N$.
The resulting node embeddings \( h_0=\{h_0^i\}_{i=0}^n \) are then passed through  \( \mathbb{N} \) stacked attention layers to obtain the final node embeddings \( h_{\mathbb{N}}= \{h_{\mathbb{N}}^i\}_{i=0}^n\) . These embeddings serve as inputs to both the node-decoder and the loc-decoder.

\subsection{Dual-decoder}
The dual-decoder architecture consists of two components: the \textit{node-decoder}, which selects the next node to visit (including both target points and the depot), and the \textit{loc-decoder}, which determines the waypoint on the boundary of the corresponding neighborhood that the agent should pass through. At each time step, the node-decoder and loc-decoder operate sequentially. Moreover, the output of the node-decoder serves as part of the input to the loc-decoder.
To enhance the loc-decoder's contextual awareness, we incorporate geometry-aware $k$-NN graph information. This augmentation enables the policy network to capture the local spatial structure around the selected neighborhood, allowing the loc-decoder to place traversal waypoints precisely.
We elaborate on the designs of the node-decoder and loc-decoder in the following subsections.

\noindent\textbf{Node-decoder.} The node-decoder generates a probability distribution for selecting the next node to visit, leveraging two primary embeddings: the node embeddings $h_\mathbb{N}$ produced by the encoder, and the context embedding \( q_t^c \). The context embedding \( q_t^c \) is constructed by combining the graph embedding $\bar{h}_\mathbb{N} = \frac{\sum_{i=0}^n h_\mathbb{N}^i}{n+1} $ with the embedding \( h_\mathbb{N}^{x_t} \) of the last visited node $x_t$, as follows,
\begin{equation}
    Q_t^c = \bar{h}_\mathbb{N}W_{Q}^{g} + h_\mathbb{N}^{x_t}W_{Q}^{l},\,
\end{equation}
where $W_{Q}^{g}$ and $ W_{Q}^{l}$ are trainable matrices. Similar to ~\cite{kool2018attention}, the node-decoder comprises an MHA layer and a compatibility layer. We regard the context embedding $Q_t^c$ as a query to attend to all nodes. To this end, we derive the keys and values from node embeddings $h_\mathbb{N}$ such that,
\begin{equation}
    K^c={h_\mathbb{N}}W_{K}^{c};\quad V^c=h_\mathbb{N}W_{V}^{c},
\end{equation}
where $W_{K}^c$ and $W_{V}^c$ are trainable parameters. Then the query $Q_t^c$, keys $K^c$, and values $V^c$ are input to the MHA layer to produce an updated context vector \( h_t^c \) as follows,
\begin{equation}
    h_t^c = \mathrm{MHA} \left(Q_t^c, K^c, V^c, M_t \right),
\end{equation}
where \( M_t \) is the attention mask applied at time step \( t \), used to assign \( -\infty \) to the attention scores of infeasible nodes. These infeasible nodes are those whose neighborhoods have already been visited in the agent's current partial tour.
This updated context vector \( h_t^c \) is subsequently processed through the compatibility layer, along with \( h_\mathbb{N} \), to compute the final probability distribution over all nodes via a masked softmax layer as follows,
\begin{equation} \label{eq:last score}
    p_\mathrm{node} = \mathrm{Softmax}\left(C\cdot\mathrm{Tanh}\left(\frac{h_\mathbb{N}h_t^c}{\sqrt{d_{k}}}\right)+M_t\right).
\end{equation}

To enhance solution diversity, we adopt the \emph{multistart} strategy from POMO~\cite{kwon2020pomo}, which generates \( n \) parallel trajectories simultaneously. All trajectories begin with the depot as the first node-action, since the agent must start from the depot. For the second node-action, we explicitly assign each trajectory a unique target point from the set \( \{1, 2, \dots, n\} \), ensuring diverse starting paths.
For subsequent time steps, node-actions are selected according to the learned probability distribution \( p_{\text{node}} \).

\noindent\textbf{Loc-Decoder.}
Given the node embeddings \( h_\mathbb{N} \) from the encoder and the selected node \( x_{t+1} \) from the node-decoder, the loc-decoder, comprising an MHA layer, a MLP, and a von Mises head, determines the boundary angle \( \psi_{t+1} \) of the waypoint in the neighborhood of the selected node. The best waypoint depends on the direction from which the agent arrives and on the neighborhoods around the selected one, since the next route segments should pass through or approach these neighborhoods. The loc-decoder therefore attends to a geometry-aware \( k \)-NN subgraph centered at the selected node and expresses this local context in a frame aligned with the arrival direction. For brevity, we write \( i = x_{t+1} \) in the following.

\textit{Local frame.} Let \( u_t = (o_i - y_t)/\|o_i - y_t\| \) be the arrival direction from the previous waypoint \( y_t \) to the center of \( \mathcal{N}(i) \), and let \( \phi_t \) be its angle. When \( y_t = o_i \), \( u_t \) is set to the first axis of the global frame. Relative positions are expressed in the local frame whose first axis is \( u_t \), i.e., they are rotated by \( R(-\phi_t) \), where \( R(\cdot) \) denotes a 2-D rotation matrix.

\textit{Geometry-aware \( k \)-NN subgraph.} Let \( \{x^{(1)}, \dots, x^{(k)}\} \) be the \( k \) nearest neighbors of \( i \) determined by the Euclidean distance between neighborhood centers. For each neighbor \( j \), we compute its offset in the local frame \( [l^x_j, l^y_j] = R(-\phi_t)(o_j - o_i) \), the center distance \( d_j = \|o_j - o_i\| \), and the boundary gap \( \delta_j = d_j - r_i - r_j \), and collect them in the geometric feature vector
\begin{equation}
    e_j = \left[\frac{l^x_j}{r_i}, \frac{l^y_j}{r_i}, d_j, \frac{d_j}{r_i}, \frac{l^x_j}{d_j}, \frac{l^y_j}{d_j}, r_j, \frac{r_j}{r_i}, \delta_j, \frac{\delta_j}{r_i}, v_j\right],
\end{equation}
where \( v_j \in \{0, 1\} \) indicates whether \( \mathcal{N}(j) \) has been visited (see Table~\ref{tab:features} in Appendix~\ref{app:features}). The quantities normalized by \( r_i \) describe each neighbor at the scale of the selected neighborhood, which is the scale at which the waypoint is placed. Each neighbor is represented by a token that concatenates its node embedding with an embedding of its geometric features, \( z_j = [h_\mathbb{N}^{j};\ \mathrm{MLP}_e(e_j)] \), where \( \mathrm{MLP}_e \) is a two-layer MLP. With \( Z_t = [z_{x^{(1)}}; \dots; z_{x^{(k)}}] \), the keys and values are computed as:
\begin{equation}
    K_t^l = Z_t W_K^l, \quad
    V_t^l = Z_t W_V^l,
\end{equation}
and the query combines the embedding of the selected node with the local context \( c_t = [\|o_i - y_t\|/r_i,\ r_i,\ \mathbb{I}(y_t \neq o_i)] \):
\begin{equation}
    Q_t^l = h_\mathbb{N}^{i} W_Q^n + c_t W_Q^c,
\end{equation}
where \( W_K^l \), \( W_V^l \), \( W_Q^n \), and \( W_Q^c \) are learnable projection matrices.

\textit{Reachability mask.} Neighbors whose boundaries are far from \( \mathcal{N}(i) \) relative to its radius have little influence on where the waypoint is placed on its boundary, and attending to them dilutes the attention over the relevant neighbors. We therefore mask a neighbor \( j \) with \( M_t^l(j) = -\infty \) if \( \delta_j > \rho\, r_i \) and \( M_t^l(j) = 0 \) otherwise, where \( \rho \) is a margin coefficient, and never mask the nearest neighbor. The output of the MHA layer is then given by:
\begin{equation}
    h_t^l = \mathrm{MHA} \left(Q_t^l, K_t^l, V_t^l, M_t^l \right).
\end{equation}

\textit{Von Mises head.} The resulting representation \( h_t^l \) is passed through a three-layer MLP that outputs a direction vector \( [m^x_t, m^y_t] \) and a scalar \( m^\omega_t \):
\begin{equation}
    [m^x_t, m^y_t, m^\omega_t] = W_f^3 \cdot g\left( W_f^2 \cdot g\left( W_f^1 h_t^l \right) \right),
\end{equation}
where \( W_f^1, W_f^2, W_f^3 \) are learnable weight matrices, and $g(\cdot)$ is the activation function. The mean direction \( \mu_t \) and the concentration \( \omega_t \) of the von Mises distribution are obtained as
\begin{equation}
\begin{split}
    \mu_t &= \operatorname{atan2}(m^y_t, m^x_t) + \phi_t, \\
    \omega_t &= \min\left(\mathrm{softplus}(m^\omega_t) + \omega_{\min},\ \omega_{\max}\right),
\end{split}
\end{equation}
where \( \operatorname{atan2}(\cdot,\cdot) \) is the two-argument arctangent, which returns the angle of the vector \( [m^x_t, m^y_t] \), \( \mathrm{softplus}(x) = \log(1 + e^{x}) \), and \( \omega_{\min} \) and \( \omega_{\max} \) are the lower and upper bounds of the concentration. The boundary angle then follows the density
\begin{equation}
    f_\theta^{\mathrm{loc}}(\psi \mid \mu_t, \omega_t) = \frac{\exp\left(\omega_t \cos(\psi - \mu_t)\right)}{2\pi I_0(\omega_t)},
\end{equation}
where \( I_0(\cdot) \) is the modified Bessel function of the first kind of order zero. The boundary angle is periodic, i.e., \( \psi \) and \( \psi+2\pi \) give the same waypoint, whereas a distribution on the interval \( [0, 2\pi] \) treats the two ends of the interval as far apart although they are adjacent on the boundary; in our preliminary experiments, a policy that sampled the angle from a Gaussian distribution on \( [0, 2\pi] \) collapsed during training. The von Mises distribution is the counterpart of the Gaussian on the circle, so its density is continuous along the whole boundary, and its entropy has a closed form, which gives the exploration term in Eq.~(\ref{eq:loss}). Predicting the direction as a 2-D vector likewise avoids the discontinuity at \( \pm\pi \) of a directly regressed angle, and adding \( \phi_t \) rotates the direction predicted in the local frame back to the global frame. The waypoint is \( y_{t+1} = y(i, \psi_{t+1}) \) according to Eq.~(\ref{eq:waypoint}).

This design offers three technical advantages. First, since the geometric features and the predicted direction are expressed in the local frame, the geometric reasoning of the loc-decoder is independent of the global orientation of the local configuration, and the network does not need to learn the same configuration separately for every arrival direction. Second, the fixed \( k \) keeps the size of the local subgraph independent of the problem size, so the loc-decoder receives inputs of the same structure on small and large instances. Third, the von Mises head places the waypoint anywhere on the boundary, which removes the discretization error of candidate-point schemes.

During training, the dual-decoder samples the next node from \( p_\theta^{\mathrm{node}} \) and the boundary angle from the von Mises distribution to promote exploration in the solution space. During testing, a greedy strategy is used, selecting the most probable node and the mean direction \( \psi_{t+1} = \mu_t \) at each step to construct the final solution. This trade-off ensures efficient exploration during training and fast inference during deployment.

\begin{algorithm}[tb]
    \vspace{-1mm}
    \caption{Customized REINFORCE for UD3RL}
    \label{alg:algorithm}
    \begin{algorithmic}[1]
        \Require Instance distribution $\mathcal{S}_{I}$; radius type distribution $\mathcal{S}_{\lambda}$; problem size distribution $\Lambda$; number of training epochs $E$; number of training instances per epoch $\mathbb{D}$; batch size $B$; number of solutions $n$
        \Ensure Trained UD3RL model $p_{\theta}$
        
        \State Initialize policy network parameters $\theta$
        \For{$e \gets 1$ to $E$}
            \State $D \gets 0$
            \While{$D < \mathbb{D}$}
                \State $\tilde{B} \gets \min(\mathbb{D}-D, B)$
                \State $\kappa \sim \textsc{SampleProblemSize}(\Lambda)$
                \State $\lambda \sim \textsc{SampleRadiusType}(\mathcal{S}_{\lambda})$
                \State $s_b \sim \textsc{SampleInstance}(\mathcal{S}_{I}), \;\forall b \in \{1, \dots, \tilde{B}\}$
                \State $\pi_b^j \sim \textsc{SampleSolution}\big(p_{\theta}(\cdot \mid s_b, \lambda, \kappa)\big), \;\forall j \in \{1,\dots,n\}, \forall b \in \{1,\dots,\tilde{B}\}$
                \State Compute gradient $\nabla J(\theta)$ according to Eq.~(\ref{eq:loss})
                \State $\theta \gets \textbf{ADAM}(\theta, \nabla J(\theta))$
                \State $D \gets D + \tilde{B}$
            \EndWhile
        \EndFor
    \end{algorithmic}
    \vspace{-1mm}
\end{algorithm}

\subsection{Objective and Optimization}
\label{sec:training}
The customized REINFORCE algorithm for UD3RL is summarized in Algorithm~\ref{alg:algorithm}. We train the policy of node selection and waypoint determination using policy gradient with a shared baseline.
Given an instance $s$, we generate multiple solutions in which the second nodes are forced to select different target points. Besides, the policy network $p_\theta$ picks an action tuple $a_t=(x_{t+1},\psi_{t+1})$ at each time step, drawing the node from the categorical distribution of the node-decoder and the boundary angle from the von Mises distribution of the loc-decoder. Then, we calculate the rewards of these solutions as the negative of the total travel distances. With the rewards, we customize REINFORCE~\cite{williams1992simple} to compute the gradient for maximizing the expected return $J(\theta) = E_{\pi \sim p_{\theta}, s \sim \mathcal{S}_{I}, \lambda \sim \mathcal{S}_{\lambda}, \kappa \sim \Lambda}R(\pi|s,\lambda, \kappa)$ such that,
\begin{equation}\label{eq:reinforce}
\begin{split}
    \triangledown J(\theta) = &E_{\pi \sim p_{\theta}, s \sim \mathcal{S}_{I}, \lambda \sim \mathcal{S}_{\lambda}, \kappa \sim \Lambda}[(R(\pi|s, \lambda, \kappa) \\
    &\,\,\,\,\,\,\,\,\,\,-\tilde{R}(s, \lambda, \kappa))\triangledown_{\theta}\log p_{\theta}(\pi|s,\lambda,\kappa)],
\end{split}    
\end{equation}
where $\tilde{R}(s, \lambda, \kappa)$ is the baseline of expected reward to reduce the variance of the sampled gradients, and $\log p_{\theta}(\pi|s,\lambda,\kappa)=\sum_{t=1}^{T}\big[\log p_\theta^{\mathrm{node}}(x_t \mid \cdot) + \log f_\theta^{\mathrm{loc}}(\psi_t \mid \cdot)\big]$ sums the log-probabilities of the node choices and the log-densities of the boundary angles. We use Monte Carlo sampling to approximate $J$ by training the policy with minibatches of instances and sampling multiple solutions for each instance. Specifically, given $B$ instances $\{s_{b}\}_{b=1}^{B}$ with radius type $\lambda$ and problem size $\kappa$ and $n$ solutions $\{\pi_{b}^{j}\}_{j=1}^{n}$ for each instance $s_{b}$, we calculate the approximate gradient as below,
\begin{equation}\label{eq:loss}
\begin{split}
    \triangledown J(\theta) \simeq \frac{1}{Bn}&\sum_{b=1}^{B}\sum_{j=1}^{n}\hat{A}_{b}^{j}\triangledown_{\theta}\log p_{\theta}(\pi_{b}^{j}|s_b, \lambda, \kappa) \\
    &+\beta\triangledown_{\theta}\bar{H},
\end{split}
\end{equation}
where $\hat{A}_{b}^{j}=\big(R(\pi_{b}^{j}|s_b, \lambda, \kappa)-\tilde{R}(s_b,\lambda, \kappa)\big)/\sigma_b$ is the advantage normalized by the standard deviation $\sigma_b$ of the $n$ rewards of instance $s_b$, and $\tilde{R}(s_b, \lambda, \kappa)\!=\!\frac{1}{n}\sum\nolimits_{j=1}^{n}R(\pi_{b}^{j}|s_b, \lambda, \kappa)$, with $n$ tours sampled by forcing different second-node choices following~\cite{kwon2020pomo}. The normalization keeps the scale of the policy gradient comparable across problem sizes and radius types, whose tour lengths differ considerably. $\bar{H}$ is the average entropy of the von Mises distributions over all waypoint decisions in the minibatch, where a distribution with concentration $\omega$ has the entropy $H(\omega)=\log\big(2\pi I_0(\omega)\big)-\omega I_1(\omega)/I_0(\omega)$, and $\beta$ is its weight. This entropy term encourages the exploration of boundary angles during training.

\noindent\textbf{Radius Type and Problem Size Sampling.} To strengthen the capability of UD3RL in solving various CETSPs, we train it with instances of different radius types and multiple sizes at the same time. Specifically, we sample a radius type $\lambda$ and a problem size $\kappa$ from predefined sets $\mathcal{S}_{\lambda}$ and $\Lambda$ respectively for each minibatch and generate training instances of the corresponding radius type and problem size for optimization as in Eq.~(\ref{eq:loss}). 

\section{Experiments} \label{sec:exp}

\subsection{Experimental Setup}
\noindent\textbf{Problems and Training.} We conduct extensive experiments to validate the effectiveness of UD3RL on CETSP across different radius configurations and problem scales. 
We train a unified UD3RL model capable of generalizing across varying problem sizes and neighborhood radius types. The training data consists of 100,000 randomly generated instances per epoch, with 2D coordinates uniformly sampled from $[0, 1]^2$. Problem sizes are uniformly drawn from $\Lambda = \{20, 40, 60, 80, 100\}$, representing the number of target nodes in each instance. For radius configurations, two types are considered: (i) constant radii, where $r = 0.1, 0.05, 0.05, 0.02,$ and $0.02$ for $n = 20, 40, 60, 80,$ and $100$, respectively; and (ii) random radii, where each radius is uniformly sampled from $[0, 0.1)$.
We train UD3RL for 1000 epochs using the Adam optimizer with learning rate $10^{-4}$ and weight decay $10^{-6}$, and reduce the learning rate by a factor of 10 after epoch 501. The batch size is set to 64. Our network architecture employs a 3-layer Transformer encoder with 8 attention heads and a hidden dimension of 128. The loc-decoder uses $k=8$ neighbors and the margin coefficient $\rho=1$, the concentration is bounded by $\omega_{\min}=0.05$ and $\omega_{\max}=30$, and the entropy weight is $\beta=10^{-3}$. 

\noindent\textbf{Baselines.} We compare UD3RL with a hybrid DRL–SOCP method and five heuristic methods. The hybrid DRL–SOCP method integrates POMO with SOCP (i.e., POMO-SOCP), where POMO determines the node visiting sequence and SOCP computes the precise waypoint within each neighborhood. The heuristic methods include the memetic algorithm MA-CETSP~\cite{lei2024effective}, which is the state-of-the-art heuristic for the CETSP, GSOA~\cite{faigl2018gsoa}, LKH~\cite{helsgaun2000effective}, OR-Tools~\cite{ortools_routing}, and a genetic algorithm (GA). The last three evaluate a visiting order by the SOCP, whose optimum is the length of the shortest tour visiting the neighborhoods in this order. LKH and OR-Tools start from TSP tours over the target centers and over the boundary distances $\max(0, \|o_i - o_j\| - r_i - r_j)$, and alternate between the SOCP and a TSP over the resulting waypoints, solved by LKH-3 or by OR-Tools with guided local search, until the tour length stops decreasing; OR-Tools has $\max(0.2, n/100)$ seconds per TSP. GA starts from 30 visiting orders, two nearest-neighbor tours and 28 random orders, each improved by 2-opt over the target centers, and evolves them by order crossover and inversion, swap, and insertion mutations with the SOCP length as fitness. Its time limit is the average time of LKH per instance of the group, with a minimum of 0.5 seconds, and the initial population is always evaluated in full. MA-CETSP and GSOA are run once per instance with their official implementations and default parameters. MA-CETSP terminates after 500 generations without improvement, and MA-CETSP (1\,s) reports the best solution that MA-CETSP holds after one second per instance.

\begin{table*}[!htbp]
  \centering
  \vspace{-3mm}
  \caption{Results on CETSP instances with different problem sizes and radius types (constant radii: neighborhood radius is fixed; random radii: neighborhood radius is uniformly sampled from a predefined range).}
  \vspace{-1mm}
  \resizebox{\textwidth}{!}{
  \renewcommand{\arraystretch}{0.8}
    \begin{tabular}{l|ccc|ccc|ccc|c}
    \toprule
    \multirow{2}[4]{*}{Method} & \multicolumn{3}{c|}{Constant radii n=20} & \multicolumn{3}{c|}{Constant radii n=50} & \multicolumn{3}{c|}{Constant radii n=100} & Constant Radii \\
\cmidrule{2-11}    \multicolumn{1}{c|}{} & \multicolumn{1}{c}{Obj.} & \multicolumn{1}{c}{Gap} & Time  & \multicolumn{1}{c}{Obj.} & \multicolumn{1}{c}{Gap} & Time  & \multicolumn{1}{c}{Obj.} & \multicolumn{1}{c}{Gap} & Time  & \multicolumn{1}{c}{Avg. Gap} \\
    \midrule
    MA-CETSP & \textbf{2.67} & \textbf{0.00\%} & 8.15m & \underline{4.12} & \underline{0.00\%} & 17.61m & 6.29 & 0.00\% & 39.49m & \underline{0.00\%} \\
    MA-CETSP (1\,s) & \underline{2.67} & \underline{0.03\%} & 1.67m & 4.13 & 0.27\% & 1.67m & 6.43 & 2.34\% & 1.67m & 0.88\% \\
    GSOA & 2.71 & 1.53\% & 0.08s & 4.23 & 2.79\% & 0.39s & 6.52 & 3.71\% & 1.54s & 2.68\% \\
    LKH & 2.71 & 1.61\% & 10.13s & 4.20 & 1.92\% & 23.59s & 6.35 & 1.02\% & 2.16m & 1.52\% \\
    ORtools & 2.71 & 1.41\% & 2.15m & 4.23 & 2.64\% & 5.34m & 6.51 & 3.51\% & 11.02m & 2.52\% \\
    GA & 2.68 & 0.22\% & 50.58s & 4.20 & 1.90\% & 1.12m & 6.52 & 3.79\% & 2.36m & 1.97\% \\
    POMO-SOCP & 2.78 & 4.15\% & 10.66s & 4.36 & 5.91\% & 17.16s & 6.80 & 8.19\% & 29.24s & 6.08\% \\
    POMO-SOCP-Aug & 2.73 & 2.28\% & 24.89s & 4.24 & 2.99\% & 48.29s & 6.62 & 5.32\% & 1.46m & 3.53\% \\
    \textbf{UD3RL} & 2.68 & 0.49\% & 0.10s & 4.19 & 1.81\% & 0.27s & 6.54 & 4.07\% & 0.68s & 2.12\% \\
    \textbf{UD3RL-Aug} & 2.68 & 0.30\% & 0.18s & 4.16 & 1.13\% & 0.76s & 6.47 & 2.89\% & 3.32s & 1.44\% \\
    \textbf{UD3RL-Aug+SOCP} & \underline{2.67} & \underline{0.03\%} & 1.37s & 4.13 & 0.40\% & 2.84s & 6.36 & 1.16\% & 6.08s & 0.53\% \\
    \textbf{UD3RL-Aug+LS (1\,s)} & \textbf{2.67} & \textbf{0.00\%} & 1.68m & \textbf{4.12} & \textbf{$-$0.02\%} & 1.69m & \underline{6.28} & \underline{$-$0.01\%} & 1.74m & \textbf{$-$0.01\%} \\
    \textbf{UD3RL-Aug+LS (10\%)} & \textbf{2.67} & \textbf{0.00\%} & 49.54s & \textbf{4.12} & \textbf{$-$0.02\%} & 1.79m & \textbf{6.28} & \textbf{$-$0.02\%} & 4.02m & \textbf{$-$0.01\%} \\
    \midrule
    \multirow{2}[4]{*}{Method} & \multicolumn{3}{c|}{Random radii n=20} & \multicolumn{3}{c|}{Random radii n=50} & \multicolumn{3}{c|}{Random radii n=100} & Random Radii \\
\cmidrule{2-11}    \multicolumn{1}{c|}{} & \multicolumn{1}{c}{Obj.} & \multicolumn{1}{c}{Gap} & Time  & \multicolumn{1}{c}{Obj.} & \multicolumn{1}{c}{Gap} & Time  & \multicolumn{1}{c}{Obj.} & \multicolumn{1}{c}{Gap} & Time  & \multicolumn{1}{c}{Avg. Gap} \\
    \midrule
    MA-CETSP & \underline{3.21} & \underline{0.00\%} & 8.38m & \underline{4.26} & \underline{0.00\%} & 15.32m & 5.23 & 0.00\% & 35.27m & 0.00\% \\
    MA-CETSP (1\,s) & 3.21 & 0.02\% & 1.67m & 4.27 & 0.12\% & 1.67m & 5.31 & 1.55\% & 1.67m & 0.56\% \\
    GSOA & 3.26 & 1.68\% & 0.08s & 4.37 & 2.57\% & 0.38s & 5.41 & 3.42\% & 1.36s & 2.56\% \\
    LKH & 3.24 & 0.85\% & 8.79s & 4.35 & 2.10\% & 24.72s & 5.51 & 5.36\% & 1.91m & 2.77\% \\
    ORtools & 3.24 & 0.83\% & 2.13m & 4.37 & 2.56\% & 5.68m & 5.64 & 7.94\% & 12.52m & 3.77\% \\
    GA & 3.22 & 0.15\% & 50.60s & 4.34 & 1.92\% & 1.14m & 5.57 & 6.58\% & 2.47m & 2.88\% \\
    POMO-SOCP & 3.38 & 5.26\% & 13.18s & 4.49 & 5.40\% & 20.74s & 5.60 & 7.13\% & 35.91s & 5.93\% \\
    POMO-SOCP-Aug & 3.26 & 1.53\% & 26.63s & 4.39 & 3.05\% & 49.67s & 5.51 & 5.40\% & 1.67m & 3.33\% \\
    \textbf{UD3RL} & 3.24 & 1.05\% & 0.09s & 4.34 & 1.76\% & 0.25s & 5.38 & 2.86\% & 0.57s & 1.89\% \\
    \textbf{UD3RL-Aug} & 3.23 & 0.67\% & 0.15s & 4.31 & 1.12\% & 0.72s & 5.32 & 1.82\% & 2.63s & 1.20\% \\
    \textbf{UD3RL-Aug+SOCP} & 3.21 & 0.09\% & 1.30s & 4.27 & 0.32\% & 2.50s & 5.27 & 0.74\% & 5.72s & 0.38\% \\
    \textbf{UD3RL-Aug+LS (1\,s)} & \textbf{3.21} & \textbf{$-$0.02\%} & 1.68m & \textbf{4.26} & \textbf{$-$0.03\%} & 1.69m & \underline{5.23} & \underline{$-$0.04\%} & 1.73m & \underline{$-$0.03\%} \\
    \textbf{UD3RL-Aug+LS (10\%)} & \textbf{3.21} & \textbf{$-$0.02\%} & 50.89s & \textbf{4.26} & \textbf{$-$0.03\%} & 1.55m & \textbf{5.22} & \textbf{$-$0.06\%} & 3.59m & \textbf{$-$0.04\%} \\
    \bottomrule
    \end{tabular}%
    }
  \label{tab:comparison}%
  \vspace{-3mm}
\end{table*}%

\noindent\textbf{Instance Augmentation.} 
We apply instance augmentation to improve the inference performance of UD3RL and POMO-SOCP. Following~\cite{kwon2020pomo}, we generate eight equivalent instances using rotation and reflection transformations that preserve the optimal tour length due to geometric symmetry: $(cx', cy') \in \{(cx, cy), (cy, cx), (cx, 1-cy), (cy, 1-cx), (1-cx, cy),$ $(1-cy, cx), (1-cx, 1-cy), (1-cy, 1-cx)\}$.
Each test instance is solved over all eight augmented versions in parallel, and the best solution is selected. Results obtained with this strategy are marked with the suffix “-Aug.”

\noindent\textbf{Post-processing.} We further report three variants that refine the tour of UD3RL-Aug. UD3RL-Aug+SOCP keeps the visiting order and re-optimizes all waypoints by the SOCP. UD3RL-Aug+LS starts a local search with 2-opt and Or-opt moves and perturbations from the tour (Appendix~\ref{app:ls}). UD3RL-Aug+LS (1\,s) runs the local search for one second per instance, and UD3RL-Aug+LS (10\%) for 10\% of the time that MA-CETSP takes on average per instance of the group.

\noindent\textbf{Metrics $\&$ Inference.} Evaluation is conducted using three metrics: objective value (Obj.), optimality gap (Gap), and runtime (Time). The Obj. represents the average traveling distance within each instance group. The Gap is defined as the percentage difference between a method’s Obj. and the Obj. of MA-CETSP; a negative Gap indicates shorter tours than those of MA-CETSP. In Tables~\ref{tab:comparison}--\ref{tab:benchmark}, the best and second-best results are shown in bold and underlined, respectively. The Time corresponds to the total runtime required for solving all instances in a group. For each configuration, we generate 100 random test instances under both constant and random radius settings with problem sizes $n \in \{20, 50, 100\}$. To assess generalization, we further evaluate larger scales $n \in \{150, 200, 300, 500\}$ as well as publicly available CETSP benchmark instances~\footnote{https://www.minlp.org/library/problem/index.php?i=65}. The depot of the synthetic test instances is at $(0.5, 0.5)$. All methods are evaluated on a server equipped with an AMD EPYC 7742 CPU and an NVIDIA A100 GPU (80 GB). We will make our dataset and implementation code publicly available upon acceptance.

\begin{table*}[!htbp]
  \centering
  \vspace{-2mm}
  \caption{Generalization results on larger problem sizes, alternate node distributions, and varying radius ranges.}
  \resizebox{\textwidth}{!}{
  \renewcommand{\arraystretch}{0.85}
    \begin{tabular}{l|ccc|ccc|ccc|ccc}
    \toprule
    \multirow{2}[4]{*}{Method} & \multicolumn{3}{c|}{n=150} & \multicolumn{3}{c|}{n=200} & \multicolumn{3}{c|}{n=300} & \multicolumn{3}{c}{n=500} \\
\cmidrule{2-13}    \multicolumn{1}{c|}{} & \multicolumn{1}{c}{Obj.} & \multicolumn{1}{c}{Gap} & \multicolumn{1}{c|}{Time} & \multicolumn{1}{c}{Obj.} & \multicolumn{1}{c}{Gap} & \multicolumn{1}{c|}{Time} & \multicolumn{1}{c}{Obj.} & \multicolumn{1}{c}{Gap} & \multicolumn{1}{c|}{Time} & \multicolumn{1}{c}{Obj.} & \multicolumn{1}{c}{Gap} & \multicolumn{1}{c}{Time} \\
    \midrule
    MA-CETSP & 5.57 & 0.00\% & 1.03h & 6.02 & 0.00\% & 1.47h & \underline{6.73} & \underline{0.00\%} & 2.46h & \underline{7.72} & \underline{0.00\%} & 4.75h \\
    MA-CETSP (1\,s) & 6.58 & 18.12\% & 1.67m & 7.35 & 22.12\% & 1.67m & 8.50 & 26.22\% & 1.67m & 10.09 & 30.75\% & 1.67m \\
    GSOA & 5.79 & 3.84\% & 2.73s & 6.31 & 4.85\% & 4.60s & 7.10 & 5.43\% & 9.63s & 8.21 & 6.43\% & 24.79s \\
    LKH & 6.01 & 7.80\% & 5.90m & 6.61 & 9.71\% & 13.19m & 7.57 & 12.41\% & 41.64m & 9.02 & 16.82\% & 2.25h \\
    ORtools & 6.26 & 12.38\% & 19.91m & 6.97 & 15.80\% & 28.29m & 8.07 & 19.89\% & 44.75m & 9.75 & 26.32\% & 1.35h \\
    GA & 6.12 & 9.81\% & 6.08m & 6.79 & 12.78\% & 13.27m & 7.84 & 16.47\% & 41.76m & 9.38 & 21.53\% & 2.25h \\
    POMO-SOCP & 6.13 & 10.00\% & 47.61s & 6.72 & 11.61\% & 1.09m & 8.05 & 19.53\% & 1.45m & 10.26 & 32.94\% & 2.25m \\
    POMO-SOCP-Aug & 6.00 & 7.67\% & 1.77m & 6.52 & 8.28\% & 2.35m & 7.62 & 13.14\% & 3.37m & 10.07 & 30.48\% & 7.40m \\
    \textbf{UD3RL} & 5.77 & 3.48\% & 1.10s & 6.30 & 4.71\% & 1.73s & 7.19 & 6.77\% & 3.70s & 8.50 & 10.20\% & 8.94s \\
    \textbf{UD3RL-Aug} & 5.71 & 2.38\% & 6.86s & 6.23 & 3.50\% & 10.74s & 7.08 & 5.19\% & 22.26s & 8.39 & 8.68\% & 55.87s \\
    \textbf{UD3RL-Aug+SOCP} & 5.63 & 1.06\% & 13.49s & 6.14 & 2.02\% & 17.51s & 6.96 & 3.37\% & 33.61s & 8.21 & 6.33\% & 1.27m \\
    \textbf{UD3RL-Aug+LS (1\,s)} & \underline{5.57} & \underline{$-$0.08\%} & 1.80m & \underline{6.02} & \underline{$-$0.03\%} & 1.87m & 6.75 & 0.27\% & 2.06m & 7.78 & 0.83\% & 2.64m \\
    \textbf{UD3RL-Aug+LS (10\%)} & \textbf{5.57} & \textbf{$-$0.13\%} & 6.31m & \textbf{6.01} & \textbf{$-$0.23\%} & 9.02m & \textbf{6.72} & \textbf{$-$0.22\%} & 15.18m & \textbf{7.70} & \textbf{$-$0.27\%} & 29.48m \\
    \midrule
    \multirow{2}[3]{*}{Method} & \multicolumn{3}{c|}{Clustered (n=100)} & \multicolumn{3}{c|}{Mixed (n=100)} & \multicolumn{3}{c|}{Small radii (n=100)} & \multicolumn{3}{c}{Large radii (n=100)} \\
\cmidrule{2-13}    \multicolumn{1}{c|}{} & \multicolumn{1}{c}{Obj.} & \multicolumn{1}{c}{Gap} & \multicolumn{1}{c|}{Time} & \multicolumn{1}{c}{Obj.} & \multicolumn{1}{c}{Gap} & \multicolumn{1}{c|}{Time} & \multicolumn{1}{c}{Obj.} & \multicolumn{1}{c}{Gap} & \multicolumn{1}{c|}{Time} & \multicolumn{1}{c}{Obj.} & \multicolumn{1}{c}{Gap} & \multicolumn{1}{c}{Time} \\
    \midrule
    MA-CETSP & 3.60 & 0.00\% & 33.67m & 4.47 & 0.00\% & 35.35m & \underline{5.81} & \underline{0.00\%} & 42.53m & \underline{3.87} & \underline{0.00\%} & 31.44m \\
    MA-CETSP (1\,s) & 3.67 & 1.97\% & 1.67m & 4.55 & 2.00\% & 1.67m & 6.06 & 4.22\% & 1.67m & 3.99 & 3.09\% & 1.67m \\
    GSOA & 3.74 & 4.06\% & 1.28s & 4.65 & 4.22\% & 1.33s & 6.02 & 3.62\% & 1.42s & 4.00 & 3.34\% & 1.10s \\
    LKH & 3.85 & 7.07\% & 3.44m & 4.76 & 6.59\% & 3.06m & 5.93 & 2.06\% & 1.62m & 4.18 & 7.98\% & 2.85m \\
    ORtools & 3.96 & 10.06\% & 13.32m & 4.89 & 9.45\% & 13.22m & 6.10 & 4.98\% & 11.23m & 4.35 & 12.44\% & 12.74m \\
    GA & 3.85 & 6.99\% & 3.66m & 4.76 & 6.56\% & 3.40m & 6.09 & 4.75\% & 2.26m & 4.11 & 6.28\% & 3.48m \\
    POMO-SOCP & 4.01 & 11.46\% & 33.05s & 4.97 & 11.30\% & 33.63s & 6.35 & 9.24\% & 34.21s & 4.23 & 9.25\% & 33.41s \\
    POMO-SOCP-Aug & 3.82 & 6.18\% & 1.29m & 4.81 & 7.72\% & 1.44m & 6.14 & 5.62\% & 1.45m & 4.17 & 7.70\% & 1.57m \\
    \textbf{UD3RL} & 3.72 & 3.53\% & 0.61s & 4.61 & 3.20\% & 0.60s & 6.03 & 3.65\% & 0.71s & 3.94 & 1.79\% & 0.64s \\
    \textbf{UD3RL-Aug} & 3.68 & 2.27\% & 2.30s & 4.56 & 2.06\% & 2.37s & 5.96 & 2.52\% & 3.36s & 3.92 & 1.14\% & 2.61s \\
    \textbf{UD3RL-Aug+SOCP} & 3.64 & 1.06\% & 6.02s & 4.51 & 0.90\% & 6.09s & 5.87 & 1.04\% & 6.17s & 3.89 & 0.42\% & 6.75s \\
    \textbf{UD3RL-Aug+LS (1\,s)} & \underline{3.59} & \underline{$-$0.10\%} & 1.72m & \underline{4.46} & \underline{$-$0.04\%} & 1.72m & \textbf{5.81} & \textbf{$-$0.03\%} & 1.74m & \textbf{3.87} & \textbf{$-$0.17\%} & 1.73m \\
    \textbf{UD3RL-Aug+LS (10\%)} & \textbf{3.59} & \textbf{$-$0.11\%} & 3.42m & \textbf{4.46} & \textbf{$-$0.06\%} & 3.59m & \textbf{5.81} & \textbf{$-$0.03\%} & 4.32m & \textbf{3.87} & \textbf{$-$0.17\%} & 3.20m \\
    \bottomrule
    \end{tabular}%
    }
  \label{tab:generalization}%
  \vspace{-3mm}
\end{table*}%

\subsection{Comparison Study} \label{subsec:compara}
Table~\ref{tab:comparison} reports the results on the six synthetic settings, where the last column gives the average gap (Avg. Gap) over the three scales. UD3RL-Aug achieves average gaps of 1.44\% and 1.20\% to MA-CETSP under constant and random radii, respectively, while requiring substantially less computation time. Among the other baselines, only LKH at $n=100$ with constant radii, where the radius of 0.02 makes the CETSP close to the TSP over the target centers, and GA at $n=20$ yield shorter tours than UD3RL-Aug. GSOA takes 0.08--1.54 seconds, comparable to UD3RL, but its tours are longer than those of UD3RL-Aug in all six settings and than those of UD3RL in five. Even without augmentation, UD3RL obtains shorter tours than POMO-SOCP-Aug in all settings. Because POMO-SOCP-Aug uses SOCP-optimal waypoints, this advantage comes from better visiting orders. The last three rows refine the tours of UD3RL-Aug. UD3RL-Aug+SOCP lowers the average gaps to 0.53\% and 0.38\% in 1.30--6.08 seconds per group. At $n=100$, its tours are shorter than those of MA-CETSP (1\,s). The gaps of both local-search variants lie between $-$0.06\% and 0.00\% in all six settings, and UD3RL-Aug+LS (1\,s) takes at most 1.74 minutes per group. Example solutions of UD3RL-Aug+LS (10\%) are shown in Appendix~\ref{app:visualization}.

\begin{figure*}[!htbp]
\centering
\includegraphics[width=7in]{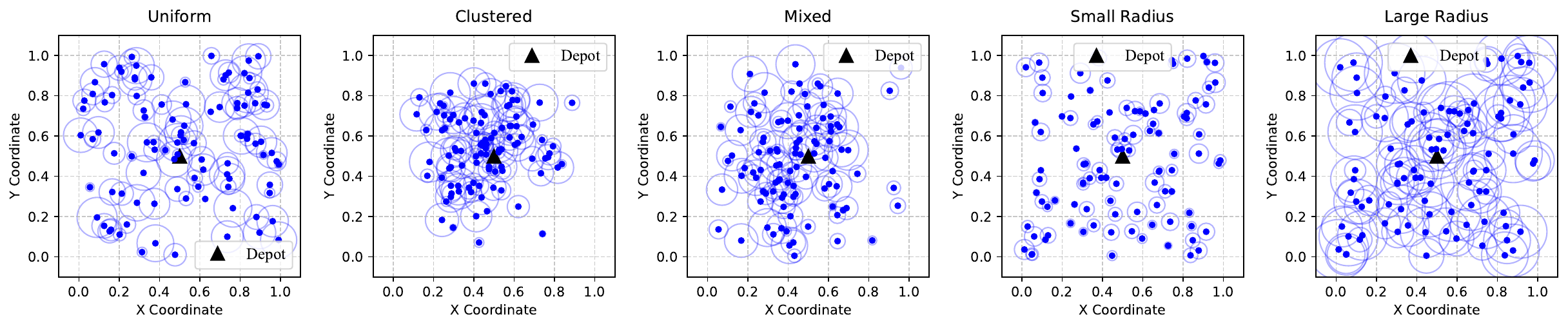}
\caption{Visualization of instances with various data distributions and radius ranges.}
\label{fig: dataset}
\end{figure*}

\subsection{Generalization Study}
We test the trained model on (1) larger sizes $n \in \{150, 200, 300, 500\}$ and (2) clustered and mixed uniform--clustered node distributions, both with radii uniformly sampled from $[0.01, 0.1]$, and (3) large radii in $[0.05, 0.15]$ and small radii in $[0.01, 0.05]$, with 100 instances per setting. Fig.~\ref{fig: dataset} illustrates the test distributions of (2) and (3) next to the uniform training distribution, and Table~\ref{tab:generalization} reports the results. On larger sizes, the gap of UD3RL-Aug to MA-CETSP grows from 2.38\% at $n=150$ to 8.68\% at $n=500$, whereas MA-CETSP (1\,s) is 18.12\%--30.75\% above MA-CETSP. UD3RL-Aug+SOCP reduces the gaps to 1.06\%--6.33\%, below all baselines except MA-CETSP. With one second of local search per instance, the gaps are at most 0.83\%, and the tours of UD3RL-Aug+LS (10\%) are 0.13\%--0.27\% shorter than those of MA-CETSP.

On clustered and mixed instances and on large radii, UD3RL-Aug stays within 2.27\%, 2.06\%, and 1.14\% of MA-CETSP. MA-CETSP (1\,s) is slightly shorter on the clustered and mixed instances (1.97\% and 2.00\%), and the gaps of all other baselines exceed 3\%. With small radii, where the CETSP is again close to the TSP over the target centers, only LKH outperforms UD3RL-Aug. In these four settings, UD3RL-Aug+SOCP has gaps of 0.42\%--1.06\%, below all baselines except MA-CETSP, and the tours of both local-search variants are 0.03\%--0.17\% shorter than those of MA-CETSP.

\subsection{Benchmark Study}
Table~\ref{tab:benchmark} reports the results on ten benchmark instances~\cite{mennell2009heuristics}, seven with constant radii and three with random radii (``rdmRad''), whose node layouts differ considerably from the uniform training distribution. MA-CETSP comes within 0.01\% of their best-known solutions~\cite{coutinho2016branch, lei2024effective}. For UD3RL, each instance is scaled into the unit square by a common factor for coordinates and radii, and the objective values are reported in the original units. Each instance is solved separately. UD3RL-Aug obtains shorter tours than GSOA, LKH, OR-Tools, and GA on six instances, and than POMO-SOCP on all ten. On rotatingDiamonds3, rotatingDiamonds5, and chaoSingleDep, whose targets lie on regular patterns, it is outperformed by LKH and GA, on the two rotatingDiamonds instances also by OR-Tools, and on rotatingDiamonds3 also by GSOA. On team6\_500, its gap of 8.95\% exceeds those of GSOA, GA, and OR-Tools: the normalized radius of 0.27 is 2.7 times the largest training radius, and each neighborhood contains 105 other target centers on average, so the eight nearest neighbors of the selected node lie inside its neighborhood and carry no information about the next route segment. UD3RL-Aug+SOCP reaches a gap of 0.00\% on five instances within 0.49 seconds, whereas its gaps on the two rotatingDiamonds instances remain 2.32\% and 6.14\%, since they stem from the visiting order. With the local search, the gap is 0.00\% on eight instances after one second and on all but team6\_500 (0.04\%) with 10\% of the time of MA-CETSP.

\begin{table*}[!htbp]
  \centering
  \vspace{-2mm}
  \caption{Results on benchmark instances.}
  \label{tab:benchmark}
  \resizebox{\textwidth}{!}{
  \renewcommand{\arraystretch}{0.95}
  \begin{tabular}{c|c|c|c|c|c|c|c|cc|cc|c|c|c}
    \toprule
    \multirow{2}[3]{*}{Instance} & \multirow{2}[3]{*}{Metrics} & \multirow{2}[3]{*}{MA-CETSP} & \multirow{2}[3]{*}{\shortstack{MA-CETSP\\(1\,s)}} & \multirow{2}[3]{*}{GSOA} & \multirow{2}[3]{*}{LKH} & \multirow{2}[3]{*}{ORtools} & \multirow{2}[3]{*}{GA} & \multicolumn{2}{c|}{POMO-SOCP} & \multicolumn{2}{c|}{\textbf{UD3RL}} & \multirow{2}[3]{*}{\shortstack{\textbf{UD3RL-Aug}\\\textbf{+SOCP}}} & \multirow{2}[3]{*}{\shortstack{\textbf{UD3RL-Aug}\\\textbf{+LS (1\,s)}}} & \multirow{2}[3]{*}{\shortstack{\textbf{UD3RL-Aug}\\\textbf{+LS (10\%)}}} \\
\cmidrule{9-12}          &       &       &       &       &       &       &       & w/o Aug & Aug & w/o Aug & Aug &       &       &       \\
    \midrule
    \multirow{3}[2]{*}{team1\_100} & Obj.   & \textbf{307.34} & 400.92 & 314.76 & 316.32 & 329.65 & 322.21 & 339.28 & 319.65 & 309.86 & 309.42 & \underline{307.35} & \textbf{307.34} & \textbf{307.34} \\
          & Gap   & \textbf{0.00\%} & 30.45\% & 2.41\% & 2.92\% & 7.26\% & 4.84\% & 10.39\% & 4.01\% & 0.82\% & 0.68\% & \underline{0.00\%} & \textbf{0.00\%} & \textbf{0.00\%} \\
          & Time  & 14.23s & 1.00s & 0.01s & 6.81s & 7.83s & 6.82s & 0.50s & 0.75s & 0.34s & 0.35s & 0.37s & 1.36s & 1.78s \\
    \midrule
    \multirow{3}[2]{*}{team6\_500} & Obj.   & \textbf{225.22} & 748.95 & 232.34 & 282.76 & 236.37 & 233.00 & 312.96 & 267.41 & 266.95 & 245.37 & 227.11 & \underline{225.30} & \underline{225.30} \\
          & Gap   & \textbf{0.00\%} & 232.55\% & 3.16\% & 25.55\% & 4.95\% & 3.46\% & 38.96\% & 18.73\% & 18.53\% & 8.95\% & 0.84\% & \underline{0.04\%} & \underline{0.04\%} \\
          & Time  & 1.25m & 1.00s & 0.18s & 2.54m & 32.09s & 2.54m & 2.11s & 9.32s & 1.71s & 2.64s & 2.90s & 3.68s & 10.19s \\
    \midrule
    \multirow{3}[2]{*}{\shortstack{team1\_100\\rdmRad}} & Obj.   & \textbf{388.54} & \textbf{388.54} & 404.70 & 413.87 & 417.89 & 404.36 & 426.08 & 397.66 & \underline{393.60} & \underline{393.60} & \textbf{388.54} & \textbf{388.54} & \textbf{388.54} \\
          & Gap   & \textbf{0.00\%} & \textbf{0.00\%} & 4.16\% & 6.52\% & 7.56\% & 4.07\% & 9.66\% & 2.35\% & \underline{1.30\%} & \underline{1.30\%} & \textbf{0.00\%} & \textbf{0.00\%} & \textbf{0.00\%} \\
          & Time  & 6.95s & 1.00s & 0.01s & 4.10s & 12.17s & 4.11s & 0.61s & 1.65s & 0.18s & 0.19s & 0.25s & 1.21s & 0.90s \\
    \midrule
    \multirow{3}[2]{*}{\shortstack{team6\_500\\rdmRad}} & Obj.   & \underline{620.91} & 751.39 & 653.84 & 762.68 & 760.86 & 786.54 & 755.12 & 719.85 & 660.88 & 648.07 & 636.73 & 622.33 & \textbf{620.90} \\
          & Gap   & \underline{0.00\%} & 21.01\% & 5.30\% & 22.83\% & 22.54\% & 26.67\% & 21.61\% & 15.93\% & 6.44\% & 4.37\% & 2.55\% & 0.23\% & \textbf{0.00\%} \\
          & Time  & 41.53s & 1.00s & 0.21s & 1.02m & 52.41s & 1.02m & 2.17s & 10.89s & 0.59s & 0.61s & 1.00s & 1.64s & 4.79s \\
    \midrule
    \multirow{3}[2]{*}{\shortstack{rotating\\Diamonds3}} & Obj.   & \textbf{380.88} & 411.01 & 384.23 & \textbf{380.88} & 383.66 & \underline{381.09} & 410.33 & 396.61 & 391.02 & 391.02 & 389.73 & \textbf{380.88} & \textbf{380.88} \\
          & Gap   & \textbf{0.00\%} & 7.91\% & 0.88\% & \textbf{0.00\%} & 0.73\% & \underline{0.05\%} & 7.73\% & 4.13\% & 2.66\% & 2.66\% & 2.32\% & \textbf{0.00\%} & \textbf{0.00\%} \\
          & Time  & 23.81s & 1.00s & 0.03s & 2.01s & 13.48s & 3.17s & 0.73s & 1.32s & 0.64s & 0.67s & 0.77s & 1.68s & 3.06s \\
    \midrule
    \multirow{3}[2]{*}{\shortstack{rotating\\Diamonds5}} & Obj.   & \textbf{1510.75} & 1596.73 & 1766.79 & 1512.09 & 1513.71 & \underline{1510.93} & 1993.74 & 1993.74 & 1612.59 & 1612.59 & 1603.53 & \textbf{1510.75} & \textbf{1510.75} \\
          & Gap   & \textbf{0.00\%} & 5.69\% & 16.95\% & 0.09\% & 0.20\% & \underline{0.01\%} & 31.97\% & 31.97\% & 6.74\% & 6.74\% & 6.14\% & \textbf{0.00\%} & \textbf{0.00\%} \\
          & Time  & 6.59m & 1.00s & 0.38s & 21.39s & 56.69s & 45.57s & 0.73s & 3.20s & 2.51s & 5.74s & 5.93s & 6.78s & 45.29s \\
    \midrule
    \multirow{3}[2]{*}{bubbles2} & Obj.   & \textbf{428.28} & \textbf{428.28} & 435.67 & 430.88 & 434.01 & 434.63 & 438.23 & 436.28 & 429.73 & \underline{429.57} & \textbf{428.28} & \textbf{428.28} & \textbf{428.28} \\
          & Gap   & \textbf{0.00\%} & \textbf{0.00\%} & 1.73\% & 0.61\% & 1.34\% & 1.48\% & 2.32\% & 1.87\% & 0.34\% & \underline{0.30\%} & \textbf{0.00\%} & \textbf{0.00\%} & \textbf{0.00\%} \\
          & Time  & 8.11s & 1.00s & 0.01s & 1.63s & 7.18s & 1.64s & 0.46s & 0.62s & 0.27s & 0.28s & 0.33s & 1.30s & 1.11s \\
    \midrule
    \multirow{3}[2]{*}{bubbles3} & Obj.   & \textbf{529.95} & 824.72 & 533.43 & 595.54 & 853.43 & 629.09 & 658.53 & 560.41 & 532.32 & 532.21 & \underline{529.96} & \textbf{529.95} & \textbf{529.95} \\
          & Gap   & \textbf{0.00\%} & 55.62\% & 0.66\% & 12.38\% & 61.04\% & 18.71\% & 24.26\% & 5.75\% & 0.45\% & 0.43\% & \underline{0.00\%} & \textbf{0.00\%} & \textbf{0.00\%} \\
          & Time  & 30.36s & 1.00s & 0.02s & 2.85s & 13.67s & 2.91s & 0.61s & 0.81s & 0.45s & 0.46s & 0.49s & 1.48s & 3.51s \\
    \midrule
    \multirow{3}[2]{*}{\shortstack{chaoSingle\\Dep}} & Obj.   & \textbf{1039.61} & 1153.40 & 1081.57 & 1041.04 & 1054.81 & 1046.98 & 1129.25 & 1098.67 & 1051.30 & 1051.11 & 1043.66 & \underline{1039.64} & \underline{1039.64} \\
          & Gap   & \textbf{0.00\%} & 10.95\% & 4.04\% & 0.14\% & 1.46\% & 0.71\% & 8.62\% & 5.68\% & 1.12\% & 1.11\% & 0.39\% & \underline{0.00\%} & \underline{0.00\%} \\
          & Time  & 26.96s & 1.00s & 0.05s & 13.48s & 15.09s & 13.51s & 0.75s & 1.54s & 0.71s & 0.74s & 0.80s & 1.75s & 3.45s \\
    \midrule
    \multirow{3}[2]{*}{\shortstack{rat195\\rdmRad}} & Obj.   & \textbf{68.22} & \textbf{68.22} & 69.57 & 72.93 & 77.24 & 71.10 & 74.89 & 70.87 & 68.60 & \underline{68.48} & \textbf{68.22} & \textbf{68.22} & \textbf{68.22} \\
          & Gap   & \textbf{0.00\%} & \textbf{0.00\%} & 1.97\% & 6.90\% & 13.21\% & 4.22\% & 9.77\% & 3.88\% & 0.55\% & \underline{0.38\%} & \textbf{0.00\%} & \textbf{0.00\%} & \textbf{0.00\%} \\
          & Time  & 5.02s & 1.00s & 0.02s & 11.63s & 14.75s & 11.69s & 0.19s & 0.63s & 0.12s & 0.13s & 0.22s & 1.14s & 0.64s \\
    \bottomrule
    \end{tabular}%
    }
    \vspace{-3mm}
\end{table*}

\begin{figure*}[htbp]
\centering
\includegraphics[width=\textwidth]{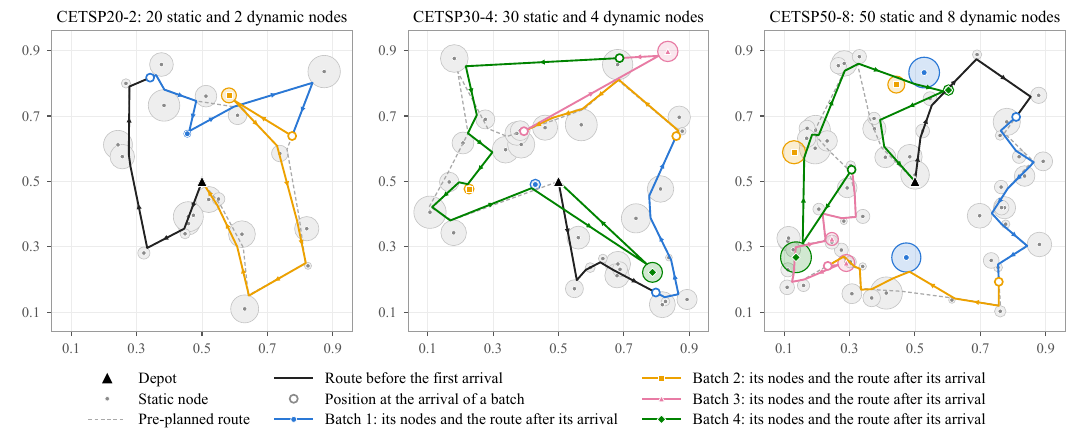}
\caption{Routes of UD3RL+MRI on dynamic CETSP instances. The dashed gray line is the pre-planned route, and the dark line is the route executed before the first batch of dynamic nodes arrives. Each batch has one color, shared by its nodes and by the route executed after its arrival; an open circle marks the position of the agent at this arrival, and the arrowheads give the travel direction.}
\label{fig: dynamic_solution}
\end{figure*}

\subsection{Performance Analysis in Dynamic Scenario}

In dynamic CETSP, new requests appear while the route is executed. Without additional training, UD3RL re-encodes the remaining and new targets, starts from the current position, and re-plans the rest of the route (Appendix~\ref{app:dynamic}).


We compare UD3RL with three insertion heuristics: cheapest insertion (CI)~\cite{wang2021xgboost, li2021learning}, modified regret insertion (MRI), and modified greedy insertion (MGI)~\cite{zhou2020two}, the last of which is used by the food delivery platform Meituan. All methods execute the route that UD3RL plans for the static nodes, so that they differ only in how the remaining route is updated when the dynamic nodes appear. The insertion heuristics insert the dynamic nodes into the remaining route. For an inserted node, its waypoint is chosen to minimize the additional travel distance between the preceding and succeeding waypoints. If the current route already intersects the node's neighborhood, the node is considered covered and requires no insertion. CI inserts the nodes in index order at their cheapest positions, MGI inserts the cheapest node--position pair at each step, and MRI inserts the node with the largest difference between its second-cheapest and cheapest insertion costs. UD3RL re-plans the remaining route, and UD3RL+MRI also repairs the previous remaining route by MRI and keeps the shorter of the two routes. Following~\cite{zhang2025neuro}, the scenarios CETSP20-2, CETSP30-4, CETSP50-8, CETSP100-20, CETSP150-30, and CETSP200-50 have 20--200 static nodes and 2--50 dynamic nodes, with 100 instances each; the depot is at $(0.5, 0.5)$, and the centers and radii of all targets are uniformly sampled from $[0.1, 0.9]^2$ and $[0.01, 0.05]$. The $m$ dynamic nodes of an instance arrive in $K=\min(4, m)$ batches of nearly equal size, and batch $j$ appears when the agent has traveled the fraction $j/(K+1)$ of the length of the initial route, so that the arrival times are the same for all methods. Static nodes crossed by the executed route are served, and a dynamic node is served only by the movement after its appearance. At each arrival, the remaining route is updated unless it already serves all pending nodes. The Obj. is the average length of the executed route, the Gap is relative to the best method, and the Time is the average time per update. As Table~\ref{tab:dynamic} shows, UD3RL+MRI obtains the shortest routes in all six scenarios, and the best insertion heuristic is 0.65\%--2.93\% longer. UD3RL alone is within 0.62\% of UD3RL+MRI in the three larger scenarios, where the insertion heuristics are at least 2.02\% longer. UD3RL re-plans the remaining route within 0.48 seconds per update and UD3RL+MRI takes up to 0.68 seconds, whereas the insertion heuristics only insert the new nodes and take 0.01--0.19 seconds. 

Fig.~\ref{fig: dynamic_solution} shows the routes of UD3RL+MRI on instances of CETSP20-2, CETSP30-4, and CETSP50-8. When the remaining route does not serve the new nodes of an arrival, the agent leaves the pre-planned route to serve them, and a node that arrives after the agent has passed its region requires a long detour.
\begin{table*}[htbp]
  \centering
  \caption{Results on dynamic CETSP instances with a common initial route and up to four arrival batches. The Time is per update.}
  \resizebox{\textwidth}{!}{
  \renewcommand{\arraystretch}{0.95}
    \begin{tabular}{c|ccc|ccc|ccc|ccc|ccc|ccc}
    \toprule
    \multirow{2}[4]{*}{Method} & \multicolumn{3}{c|}{CETSP20-2} & \multicolumn{3}{c|}{CETSP30-4} & \multicolumn{3}{c|}{CETSP50-8} & \multicolumn{3}{c|}{CETSP100-20} & \multicolumn{3}{c|}{CETSP150-30} & \multicolumn{3}{c}{CETSP200-50} \\
\cmidrule{2-19}          & \multicolumn{1}{c}{Obj} & \multicolumn{1}{c}{Gap} & \multicolumn{1}{c|}{Time} & \multicolumn{1}{c}{Obj} & \multicolumn{1}{c}{Gap} & \multicolumn{1}{c|}{Time} & \multicolumn{1}{c}{Obj} & \multicolumn{1}{c}{Gap} & \multicolumn{1}{c|}{Time} & \multicolumn{1}{c}{Obj} & \multicolumn{1}{c}{Gap} & \multicolumn{1}{c|}{Time} & \multicolumn{1}{c}{Obj} & \multicolumn{1}{c}{Gap} & \multicolumn{1}{c|}{Time} & \multicolumn{1}{c}{Obj} & \multicolumn{1}{c}{Gap} & \multicolumn{1}{c}{Time} \\
    \midrule
    CI & 3.11 & 0.65\% & 0.01s & 3.86 & 0.78\% & 0.01s & 4.80 & 1.05\% & 0.02s & 6.58 & 2.49\% & 0.06s & 7.69 & 3.36\% & 0.11s & 8.77 & 2.93\% & 0.19s \\
    MRI & 3.11 & 0.65\% & 0.01s & 3.86 & 0.78\% & 0.01s & 4.79 & 0.84\% & 0.02s & 6.55 & 2.02\% & 0.06s & 7.62 & 2.42\% & 0.11s & 8.77 & 2.93\% & 0.19s \\
    MGI & 3.11 & 0.65\% & 0.01s & 3.86 & 0.78\% & 0.01s & 4.79 & 0.84\% & 0.02s & 6.56 & 2.18\% & 0.06s & 7.71 & 3.63\% & 0.11s & 8.85 & 3.87\% & 0.19s \\
    \textbf{UD3RL} & 3.13 & 1.29\% & 0.04s & 3.87 & 1.04\% & 0.06s & 4.81 & 1.26\% & 0.11s & 6.46 & 0.62\% & 0.23s & 7.46 & 0.27\% & 0.34s & 8.54 & 0.23\% & 0.48s \\
    \textbf{UD3RL+MRI} & \textbf{3.09} & \textbf{0.00\%} & 0.05s & \textbf{3.83} & \textbf{0.00\%} & 0.07s & \textbf{4.75} & \textbf{0.00\%} & 0.13s & \textbf{6.42} & \textbf{0.00\%} & 0.28s & \textbf{7.44} & \textbf{0.00\%} & 0.45s & \textbf{8.52} & \textbf{0.00\%} & 0.68s \\
    \bottomrule
    \end{tabular}%
    }
  \label{tab:dynamic}%
  \vspace{-3mm}
\end{table*}%

\subsection{Ablation Study} \label{subsec:ablation}

To assess the key components, we compare UD3RL with two variants trained with the same settings: (i) UD3RL w/o $k$-NN, whose loc-decoder removes the subgraph; and (ii) UD3RL w/o adapted encoder, which uses the original encoder (i.e., post-instance normalizations and standard GELU-activated FF layers). As Table~\ref{tab:ablation} shows, where the Gap is relative to the best variant, UD3RL obtains the shortest tours in four of th e six settings and the same Obj. as UD3RL w/o $k$-NN at $n=20$ and $n=50$ with random radii. Removing the subgraph raises the gap by up to 0.56\%, and the variant without the adapted encoder has gaps of 0.93\%--2.23\%. 

\begin{table}[htbp]
  \centering
  \caption{Results of our ablation study.}
  \resizebox{\columnwidth}{!}{
  \renewcommand{\arraystretch}{0.85}
    \begin{tabular}{c|c|p{3em}|p{6.1em}<{\centering}|p{6.1em}<{\centering}|p{6.1em}<{\centering}}
    \toprule
    \multicolumn{2}{c|}{Instances} & \multicolumn{1}{c|}{Metrics} & UD3RL & UD3RL w/o $k$-NN & UD3RL w/o adapted encoder \\
    \midrule
    \multirow{9}[6]{*}{\begin{sideways}Constant Radii\end{sideways}} & \multicolumn{1}{c|}{\multirow{3}[2]{*}{n=20}} & Obj & \textbf{2.68} & 2.69 & 2.71 \\
          &       & Gap & \textbf{0.00\%} & 0.37\% & 1.12\% \\
          &       & Time & 0.10s & 0.05s & 0.07s \\
\cmidrule{2-6}          & \multicolumn{1}{c|}{\multirow{3}[2]{*}{n=50}} & Obj & \textbf{4.19} & 4.20 & 4.25 \\
          &       & Gap & \textbf{0.00\%} & 0.24\% & 1.43\% \\
          &       & Time & 0.27s & 0.15s & 0.18s \\
\cmidrule{2-6}          & \multicolumn{1}{c|}{\multirow{3}[2]{*}{n=100}} & Obj & \textbf{6.54} & 6.56 & 6.64 \\
          &       & Gap & \textbf{0.00\%} & 0.31\% & 1.53\% \\
          &       & Time & 0.68s & 0.32s & 0.59s \\
    \midrule
    \multirow{9}[6]{*}{\begin{sideways}Random Radii\end{sideways}} & \multicolumn{1}{c|}{\multirow{3}[2]{*}{n=20}} & Obj & \textbf{3.24} & \textbf{3.24} & 3.27 \\
          &       & Gap & \textbf{0.00\%} & \textbf{0.00\%} & 0.93\% \\
          &       & Time & 0.09s & 0.05s & 0.08s \\
\cmidrule{2-6}          & \multicolumn{1}{c|}{\multirow{3}[2]{*}{n=50}} & Obj & \textbf{4.34} & \textbf{4.34} & 4.39 \\
          &       & Gap & \textbf{0.00\%} & \textbf{0.00\%} & 1.15\% \\
          &       & Time & 0.25s & 0.13s & 0.18s \\
\cmidrule{2-6}          & \multicolumn{1}{c|}{\multirow{3}[2]{*}{n=100}} & Obj & \textbf{5.38} & 5.41 & 5.50 \\
          &       & Gap & \textbf{0.00\%} & 0.56\% & 2.23\% \\
          &       & Time & 0.57s & 0.28s & 0.49s \\
    \bottomrule
    \end{tabular}%
    }
  \label{tab:ablation}%
  \vspace{-3mm}
\end{table}%

\subsection{Continuous versus Discretized Waypoints} \label{subsec:cont_vs_disc}

We compare UD3RL with UD3RL w/ vanilla $k$-NN, which attends to the node embeddings of the $\lfloor (n+1)/4 \rfloor-1$ nearest neighbors without local frame, and with two discretized variants, which replace the von Mises head of UD3RL w/ vanilla $k$-NN with a softmax over $\gamma \in \{16, 64\}$ candidate points placed by PDS (Appendix~\ref{app:pds}). All variants are trained with the same data, optimizer, and schedule. The waypoint gap is the relative difference between the tour length of a method and the SOCP-optimal tour length for the same visiting order. As Table~\ref{tab:continuous} shows, UD3RL obtains the shortest tours in all settings, significantly shorter than those of both discretized variants, and UD3RL w/ vanilla $k$-NN, which uses the same subgraph as the discretized variants, also outperforms them. The waypoint gap of UD3RL is 0.41\%--1.81\%, against 1.51\%--3.60\% for the discretized variants, and refining the discretization from $\gamma=16$ to $\gamma=64$ increases the waypoint gap in all settings and the tour length in all but one, where the Obj. is the same, as the larger action space makes waypoint selection harder to learn.

\begin{table}[htbp]
  \centering
  \caption{Comparison of continuous and discretized waypoint selection. The waypoint gap is the relative difference between the tour length of a method and the SOCP-optimal tour length for the same visiting order.}
  \resizebox{\columnwidth}{!}{
  \renewcommand{\arraystretch}{0.9}
    \begin{tabular}{l|ccc|ccc}
    \toprule
    \multirow{2}[2]{*}{Method} & \multicolumn{3}{c|}{Constant radii} & \multicolumn{3}{c}{Random radii} \\
\cmidrule{2-7}          & n=20 & n=50 & n=100 & n=20 & n=50 & n=100 \\
    \midrule
    \multicolumn{7}{c}{Obj.} \\
    \midrule
    Discrete ($\gamma=16$) & 2.74 & 4.27 & 6.66 & 3.27 & 4.40 & 5.52 \\
    Discrete ($\gamma=64$) & 2.75 & 4.29 & 6.66 & 3.28 & 4.42 & 5.54 \\
    UD3RL w/ vanilla $k$-NN & 2.70 & 4.21 & 6.55 & 3.25 & 4.35 & 5.41 \\
    \textbf{UD3RL} & \textbf{2.68} & \textbf{4.19} & \textbf{6.54} & \textbf{3.24} & \textbf{4.34} & \textbf{5.38} \\
    \midrule
    \multicolumn{7}{c}{Waypoint gap} \\
    \midrule
    Discrete ($\gamma=16$) & 2.26\% & 2.45\% & 2.62\% & 1.51\% & 2.02\% & 2.96\% \\
    Discrete ($\gamma=64$) & 2.46\% & 3.02\% & 3.04\% & 1.66\% & 2.54\% & 3.60\% \\
    UD3RL w/ vanilla $k$-NN & 0.89\% & 1.33\% & 1.80\% & 0.77\% & 1.14\% & 1.66\% \\
    \textbf{UD3RL} & 0.41\% & 0.98\% & 1.81\% & 0.84\% & 0.97\% & 1.25\% \\
    \bottomrule
    \end{tabular}%
    }
  \label{tab:continuous}%
  \vspace{-3mm}
\end{table}%

\section{Conclusion} \label{sec:conclude}
This paper presents UD3RL, a unified dual-decoder DRL framework for efficient close-enough routing. The CETSP involves coupled discrete node-ordering and continuous waypoint-selection decisions, and UD3RL handles both directly with a hybrid discrete--continuous policy. The node-decoder determines the next target neighborhood to visit, while the loc-decoder outputs a von Mises distribution over the boundary angle of the waypoint in that neighborhood. A geometry-aware $k$-NN subgraph interaction strategy further enhances local spatial reasoning during waypoint selection. Experiments show that UD3RL achieves competitive solution quality with substantially lower computation time than strong CETSP baselines, while further refinement with lightweight local search matches or outperforms MA-CETSP across the synthetic test settings. UD3RL also generalizes well across problem scales, node distributions, and radius settings, and effectively adapts to dynamic requests without retraining. These results highlight the potential of UD3RL for efficient close-enough routing in intelligent transportation applications, including automatic meter reading, UAV-based inspection and data collection, and dynamic delivery routing. Future work will extend UD3RL to road-network travel distances, extremely large-scale instances, and cooperative close-enough routing with multiple vehicles.

\section*{Acknowledgments}
We used ChatGPT to assist in the linguistic refinement of the Introduction and Conclusion sections. 
The generated text was critically reviewed and revised by the authors to ensure alignment with the research findings and academic standards.

\bibliographystyle{IEEEtran}
\bibliography{main}

\input{appendix}



\end{document}

%% file: appendix.tex
\appendices
\section{Discretized Variant with PDS}
\label{app:pds}
\begin{figure*}[!htbp]
\centering
\vspace{-1mm}
\includegraphics[width=0.88\textwidth]{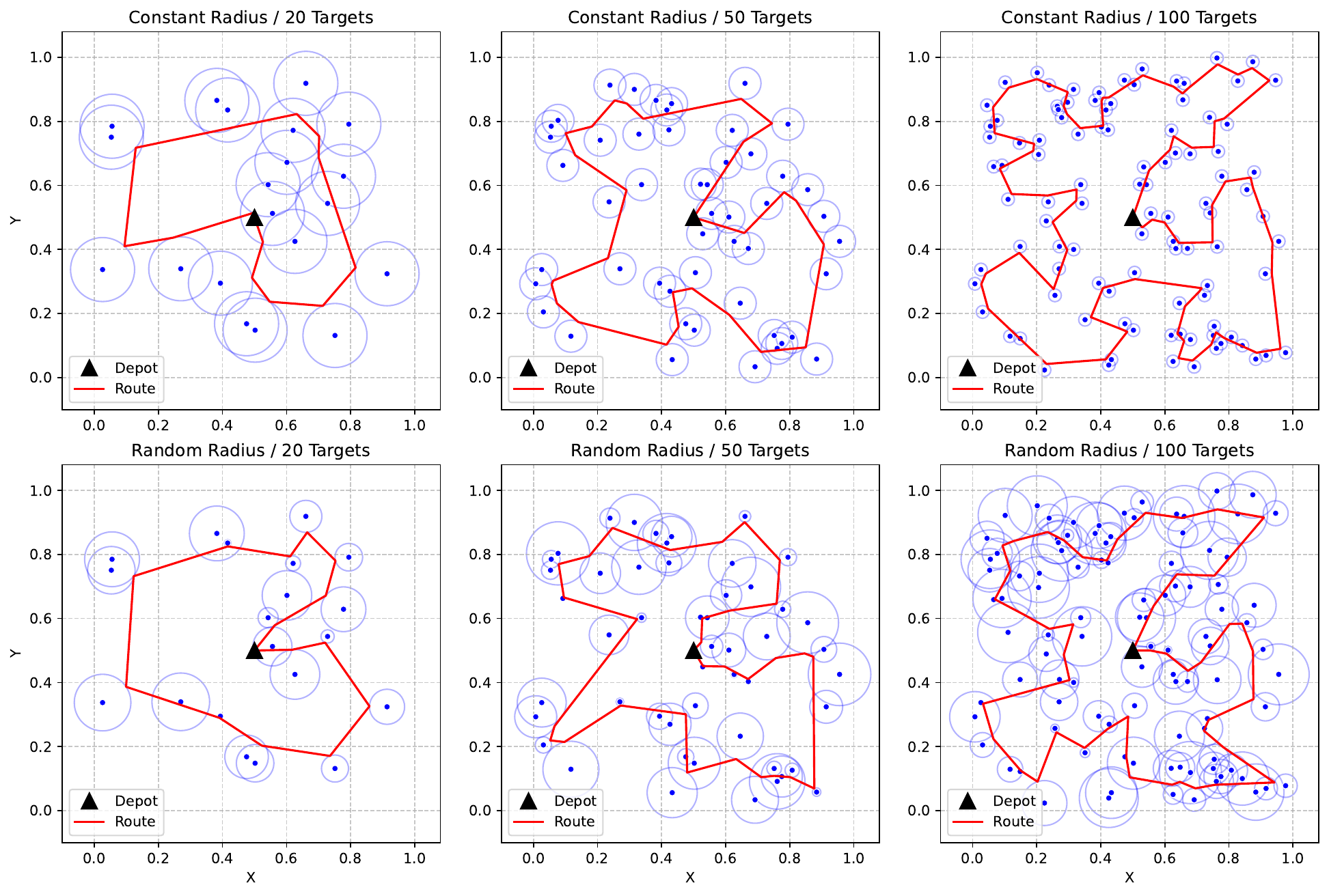}
\caption{Visualization of CETSP solutions obtained by UD3RL-Aug+LS (10\%) across different problem scales. The top row shows results under the constant radius configuration, while the bottom row corresponds to the random radius configuration.}
\label{fig: compara routes}
\vspace{-3mm}
\end{figure*}

\begin{figure}[t]
\vspace{-3mm}
\centering
\includegraphics[width=0.4\linewidth]{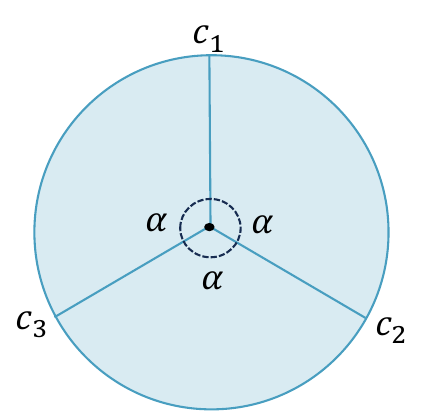}
\caption{PDS for $\gamma = 3$ with $\hat{N}(i) = \{c1, c2, c3\}$ and $\alpha = 120\degree$.}
\label{fig:circle}
\vspace{-3mm}
\end{figure}

PDS places discretization points uniformly along the circumference of each target neighborhood. Specifically, let \( \gamma \) denote the number of points used to discretize each neighborhood \( \mathcal{N}(i)\), denoted by \( \hat{\mathcal{N}}(i)=\{c_1, \cdots, c_\gamma\}\). Then, the PDS divides the circumference \( C_i \) of each neighborhood into \( \gamma \) equal circular arcs with central angle $\alpha=\frac{360^\circ}{\gamma}$, and places discretization points at the endpoints of these arcs, as shown in Fig.~\ref{fig:circle}. 
For the classical TSP with $n$ cities, the time complexity is $\mathcal{O}(n!)$. In contrast, after applying PDS to the CETSP, the solution space expands to all sequences of discretization points, resulting in a time complexity of $\mathcal{O}(n!\gamma^n)$, which is significantly higher than that of the TSP due to the added combinatorial burden from continuous-to-discrete transformation.

The discretized variants of UD3RL in Section~\ref{subsec:cont_vs_disc} restrict the waypoint in \( \mathcal{N}(i) \) to \( \hat{\mathcal{N}}(i) \), i.e., to the angles \( \psi \in \{0, \alpha, \dots, (\gamma-1)\alpha\} \) in Eq.~(\ref{eq:waypoint}). Their loc-decoder uses the vanilla $k$-NN subgraph interaction of Section~\ref{subsec:cont_vs_disc}, and its MLP outputs \( \gamma \) logits followed by a softmax layer,
\begin{equation}
    p_{\text{loc}} = \mathrm{Softmax}\left( W_f^3 \cdot g\left( W_f^2 \cdot g\left( W_f^1 h_t^l \right) \right) \right),
\end{equation}
from which the waypoint is sampled during training and selected greedily during testing.

\section{Geometric Features of the Loc-Decoder}
\label{app:features}
Table~\ref{tab:features} lists the geometric features of a neighbor \( j \) of the selected neighborhood \( \mathcal{N}(i) \) used by the geometry-aware \( k \)-NN subgraph interaction.

\begin{table}[htbp]
  \centering
  \caption{Geometric features of a neighbor \( j \) of the selected neighborhood \( \mathcal{N}(i) \).}
  \label{tab:features}
  \renewcommand{\arraystretch}{1.1}
  \begin{tabular}{c|p{5.4cm}}
    \toprule
    Feature & Description \\
    \midrule
    \( l^x_j/r_i,\ l^y_j/r_i \) & Offset of \( o_j \) from \( o_i \) in the local frame, in units of \( r_i \) \\
    \( d_j,\ d_j/r_i \) & Center distance, absolute and in units of \( r_i \) \\
    \( l^x_j/d_j,\ l^y_j/d_j \) & Direction of \( o_j \) in the local frame \\
    \( r_j,\ r_j/r_i \) & Radius of \( \mathcal{N}(j) \), absolute and relative to \( r_i \) \\
    \( \delta_j,\ \delta_j/r_i \) & Boundary gap \( d_j - r_i - r_j \), absolute and in units of \( r_i \) \\
    \( v_j \) & Whether \( \mathcal{N}(j) \) has been visited \\
    \bottomrule
  \end{tabular}
\end{table}

\section{Details of Attention Layer}
\label{app:attention}
Each attention layer includes a multi-head attention (MHA) sublayer and a feed-forward (FF) sublayer.

\noindent\textbf{MHA Sublayer.} The MHA sublayer captures dependencies between different nodes in the input sequence. Given the node embeddings $h_{l-1}=\{h_{l-1}^{i}\}_{i=0}^{n}$ as inputs to the $l$-th attention layer, we first employ RMS normalization to obtain pre-norm node embeddings: ${h_{l-1}}' = \mathrm{RMSNorm}(h_{l-1})$.

We begin by linearly projecting the pre-normalized node embeddings \( {h_{l-1}}' \) into \emph{queries} \( Q \), \emph{keys} \( K \), and \emph{values} \( V \), using \( H \) distinct subspaces (attention heads). Each projection is performed using learned weight matrices \( W_j^Q \), \( W_j^K \), and \( W_j^V \), such that for each head \( j = 1, \ldots, H \):
\[
Q_j = {h_{l-1}}' W_j^Q, \quad K_j = {h_{l-1}}' W_j^K, \quad V_j = {h_{l-1}}' W_j^V.
\]
The attention weights for each head are computed using scaled dot-product attention, followed by a softmax operation:
\begin{equation}
A_j = \mathrm{Softmax}\left( \frac{Q_j K_j^\top}{\sqrt{d_k}} + M\right),
\label{eq:attn}
\end{equation}
where \( d_k \) is the dimensionality of the key vectors, used as a scaling factor to stabilize gradients, and $M$ is an optional attention mask. This mask can be used to prevent attending to infeasible actions by setting corresponding entries to \( -\infty \). Note that the attention mask is not applied during the encoder. The output of each attention head is then computed as the weighted sum of the projected values: $Z_j = A_j V_j$.
The outputs from all heads are concatenated and passed through a final linear projection using a learned matrix \( W^O \), producing the final output of the MHA module:
\begin{equation}
\mathrm{MHA}(Q, K, V) = \mathrm{Concat}(Z_1, \ldots, Z_H) W^O.
\label{eq:mha}
\end{equation}
Finally, the output of the MHA sublayer is obtained by applying a residual connection that adds the original input \( h_{l-1} \) to the MHA output:
\begin{equation}
\hat{h}_{l-1} = h_{l-1} + \mathrm{MHA}(Q, K, V).
\end{equation}

\noindent\textbf{FF Sublayer.}  
The embeddings \( \hat{h}_{l-1} \) from the MHA sublayer are passed into the feed-forward (FF) sublayer, which consists of RMS normalization, a SiLU-activated gated MLP~\cite{shazeer2020glu}, and a residual connection. The computations are defined as follows:
\begin{equation}
    \hat{h}'_{l-1} = \mathrm{RMSNorm}(\hat{h}_{l-1}),
\end{equation}
\begin{equation}
    h^{f}_{l-1} = W_3 \left( \mathrm{SiLU}\left(W_1 \hat{h}'_{l-1}\right) \odot W_2 \hat{h}'_{l-1} \right),
\end{equation}
\begin{equation}
    h_{l} = \hat{h}_{l-1} + h^{f}_{l-1},
\end{equation}
where \( \odot \) denotes element-wise multiplication, \( \mathrm{SiLU} \) is the Sigmoid Linear Unit (also known as Swish), and \( W_1, W_2, W_3 \) are learnable weight matrices.

We stack \( \mathbb{N} \) attention layers (with \( \mathbb{N} = 3 \) in our implementation) in the encoder, each following the same architecture but with independent parameters. 

\section{Additional Experimental Results}
\label{app:visualization}

The example routes of UD3RL-Aug+LS (10\%) in Fig.~\ref{fig: compara routes} pass through the target neighborhoods with compact trajectories.

\begin{figure}[!htbp]
\centering
\includegraphics[width=0.72\columnwidth]{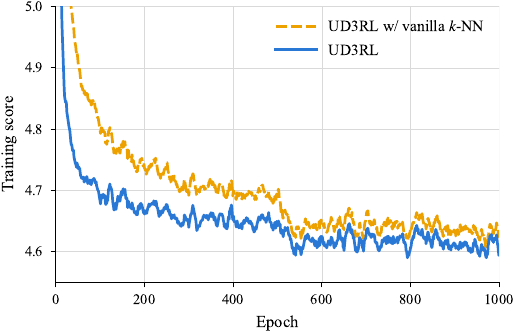}
\caption{Training scores of UD3RL and UD3RL w/ vanilla $k$-NN, smoothed by a moving average over 10 epochs. The training score is the average length of the best of the $n$ sampled tours over the training instances of an epoch.}
\label{fig:training}
\end{figure}

Fig.~\ref{fig:training} compares the training curves of UD3RL and UD3RL w/ vanilla $k$-NN, which replaces the geometry-aware subgraph with the vanilla $k$-NN subgraph of Section~\ref{subsec:cont_vs_disc}. UD3RL reaches a smoothed training score of 4.70 after 95 epochs, against 279 epochs for UD3RL w/ vanilla $k$-NN, and ends with the lower score.

\section{Dynamic Replanning Algorithm}
\label{app:dynamic}

Algorithm~\ref{alg:dynamic solve} gives the process for solving dynamic CETSP. The trained model first generates an initial route for the static nodes, and the visited-node set is updated while the route is executed. When dynamic nodes appear and the current route does not cover them, UD3RL re-encodes the unvisited static nodes and the dynamic nodes and generates a new route without retraining. The current position enters the re-encoded instance as a node with zero radius, and this node and the depot are masked at the start, so that every rollout starts from the current position, serves all remaining nodes, and returns to the depot. UD3RL+MRI additionally inserts the uncovered nodes into the previous remaining route by MRI and executes the shorter of the two routes.

\begin{algorithm}[H]
    \caption{Dynamic CETSP Solving using UD3RL}
    \label{alg:dynamic solve}
    \begin{algorithmic}[1]
        \Require Static nodes $N=\{0,1,\ldots,n\}$ with coordinates and radii; trained UD3RL model $p_{\theta}$
        \Ensure Complete solution $\pi'$ visiting all static and dynamic nodes
        \State Initialize $\mathrm{V_{nodes}} \gets \emptyset$
        \State Obtain initial solution $\pi_0$ and visiting tour $\tau$ from the trained UD3RL model
        \While{the agent has not returned to the depot}
            \State Move along the current tour; update $\mathrm{V_{nodes}}$
            \If{dynamic nodes $D$ appear}
                \State Check whether the remaining tour covers the nodes in $D$
                \If{any uncovered nodes exist}
                    \State Re-encode unvisited static and dynamic nodes $N'$
                    \State Set current state $S_t$ with the current position
                    \State Update $\pi'$ and $\tau'$ using the trained UD3RL model
                \EndIf
            \EndIf
        \EndWhile
    \end{algorithmic}
\end{algorithm}

\section{Local Search of UD3RL-Aug+LS}
\label{app:ls}

The local search represents a solution by the visiting sequence of all targets and a status of each target: an anchor has a waypoint in its neighborhood, and a passive target is served on the way by the segment between the two neighboring anchors. The initial sequence is the visiting order of UD3RL-Aug, with every target covered on the way placed at the first segment crossing its neighborhood. For fixed neighboring anchors, the optimal waypoint of an anchor lies on the segment between them if it crosses the neighborhood, and otherwise at the boundary point minimizing the sum of distances to both (Newton's method). After each change, a passive target that is no longer crossed becomes an anchor, and an anchor whose neighborhood is crossed by the segment between its neighbors becomes passive. The sequence is modified by 2-opt, which reverses a part of it, and by Or-opt, which moves up to three consecutive anchors to another position in either orientation. Candidate moves are restricted to the 10 nearest neighbors of each target, and a move is kept only if the tour is shorter after the waypoint update. When no move shortens the tour, a double-bridge perturbation exchanges two adjacent blocks of the sequence, the local search is repeated around the changed positions, and the result is accepted if it is shorter. After $\max(25, 0.1n)$ perturbations without improvement, the search continues from the best solution with four to eight perturbations at once.